\documentclass{article}
\usepackage{iclr2027_conference,times}
\iclrfinalcopy 
\date{}

\usepackage{amsmath,amsfonts,bm}

\def\1{\bm{1}}

\DeclareMathAlphabet{\mathsfit}{\encodingdefault}{\sfdefault}{m}{sl}
\SetMathAlphabet{\mathsfit}{bold}{\encodingdefault}{\sfdefault}{bx}{n}

\usepackage{url}
\usepackage{amsmath,amssymb,amsthm}
\usepackage{booktabs}
\usepackage{graphicx}
\usepackage{algorithm}
\usepackage{algpseudocode}
\usepackage{multirow}
\usepackage{tikz}
\usepackage{tabularx}
\usepackage{placeins}
\usepackage{array}
\usetikzlibrary{arrows.meta,positioning,shapes.geometric,fit,backgrounds}

\newtheorem{proposition}{Proposition}
\newtheorem{assumption}{Assumption}
\newtheorem{corollary}{Corollary}
\newtheorem{remark}{Remark}

\newcommand{\attr}{\mathsf{attr}}
\newcommand{\selferr}{\textsc{Self\_Error}}
\newcommand{\external}{\textsc{External}}
\newcommand{\unknown}{\textsc{Unknown}}
\newcommand{\falsealarm}{\textsc{False\_Alarm}}

\newcommand{\Mset}{\mathcal{M}}
\newcommand{\Aset}{\mathcal{A}}
\newcommand{\cre}{c_{\mathrm{re}}}
\newcommand{\crb}{c_{\mathrm{rb}}}

\title{AquaMend: Minimal Re-probing and Conditional Rollback for Latent-Belief Failures in Embodied Agents}

\usepackage{hyperref}
\newcommand{\affmark}[1]{\raisebox{.5ex}{\fontsize{7}{8}\selectfont #1}}
\author{\begin{minipage}{\textwidth}
\centering\normalfont\fontsize{10}{12}\selectfont
Yufan Liu\affmark{1,*}, Shang Luo\affmark{2,*}, Yang Liu\affmark{2,*}, Haoxuan Jia\affmark{5,6}\\[3pt]
Feiyu Han\affmark{7}, Qian Li\affmark{3}, Chen Li\affmark{4}, Yingguang Yang\affmark{2}\\[3pt]
Chongyang Zhang\affmark{5}, Hao Zheng\affmark{5}, Kefu Xu\affmark{2}, Bin Chong\affmark{2,\textdagger}\\[7pt]
\fontsize{9}{11}\selectfont
\affmark{1}University College London\quad
\affmark{2}Peking University\\
\affmark{3}Beijing University of Posts and Telecommunications\\
\affmark{4}University of Leeds\quad\affmark{5}Fullive-AI\\
\affmark{6}Nanyang Technological University\\
\affmark{7}University of Chinese Academy of Sciences\\[5pt]
\affmark{*}Equal contribution.\quad
\affmark{\textdagger}Corresponding author: \texttt{chongbin@pku.edu.cn}\\[5pt]
\texttt{ucemicz@ucl.ac.uk}; \texttt{liuyang2020@amss.ac.cn}\\
\texttt{zrms2172@leeds.ac.uk}; \texttt{dao@mail.ustc.edu.cn}\\
\texttt{zcyforwork1017@gmail.com}; \texttt{haozheng1750@gmail.com}\\
\texttt{xukefu@pku.edu.cn}
\end{minipage}}
\hypersetup{
  pdftitle={AquaMend: Minimal Re-probing and Conditional Rollback for Latent-Belief Failures in Embodied Agents},
  pdfauthor={Yufan Liu, Shang Luo, Yang Liu, Haoxuan Jia, Feiyu Han, Qian Li, Chen Li, Yingguang Yang, Chongyang Zhang, Hao Zheng, Kefu Xu, Bin Chong}
}

\makeatletter
\renewcommand{\@maketitle}{%
  {\LARGE\scshape\@title\par}%
  \vspace{10pt}%
  \noindent\@author\par
  \vspace{14pt}%
}
\makeatother

\providecommand{\revisioncolor}{}
\providecommand{\rev}[1]{{\revisioncolor #1}}

\begin{document}
\maketitle
\lhead{Preprint} 
\begin{abstract}
Physical changes or sensing errors can invalidate embodied agents' task-relevant beliefs.
AquaMend compares re-probing, rollback, and supported continuation on a probe--belief--action graph under an expected-loss objective covering sensing, physical recovery, and uncorrected failures.
A joint posterior guides a one-step policy with conditional detection-power screening.
The per-belief three-way optimum requires independence, separability, and fully resolving probes; the general policy has no global optimality guarantee.
Across 32 paired scenarios in a self-constructed simulation benchmark, AquaMend recovers in 28/32 cases and reduces mean complete loss by 21.6\% versus restart.
Its paired loss difference from decision-theoretic troubleshooting (DTT) is not statistically significant after Holm correction.
Against the all-candidate ablation, online decision time decreases by 12.3\% overall but increases by 3.4\% in the uncovered late stage.
\end{abstract}

\suppressfloats[t]
\begin{figure}[t]
\centering
\includegraphics[width=\linewidth]{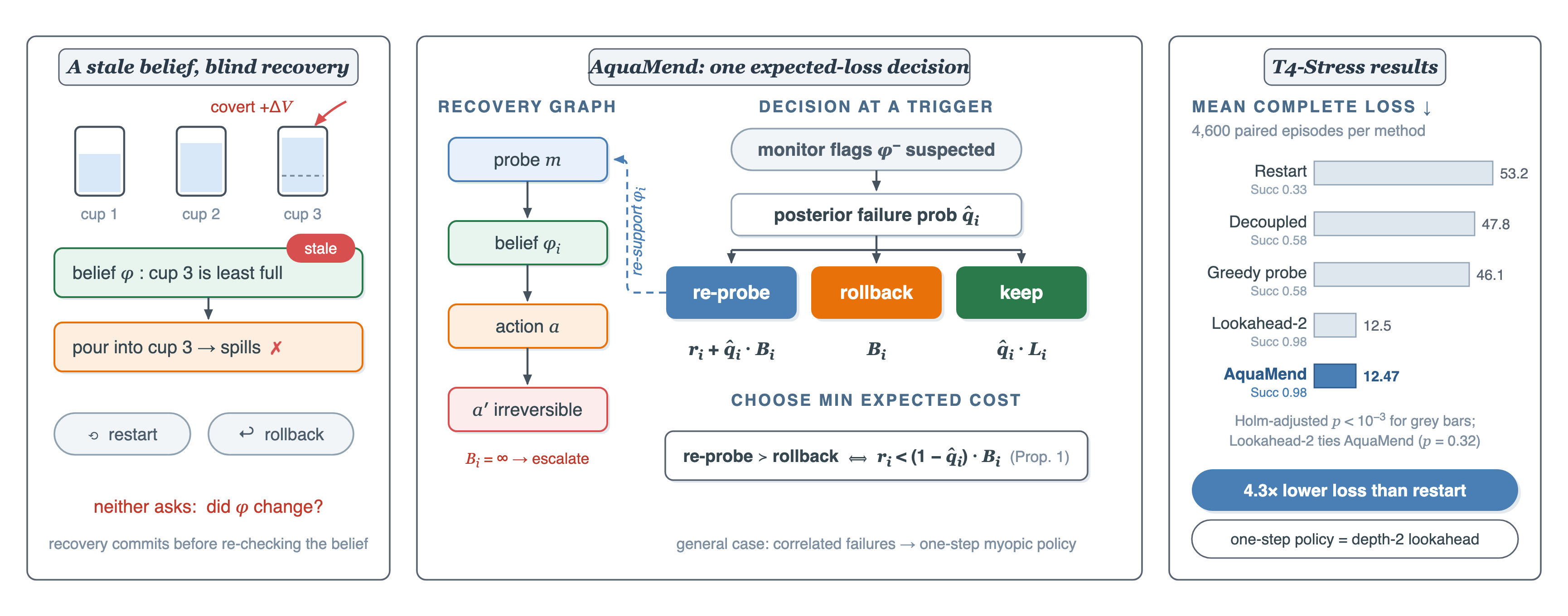}
\par\vspace{-2pt}
\caption{\textbf{AquaMend overview.}
\textbf{(a)} The restart and rollback-only baselines considered here can commit to recovery before re-checking a suspected belief.
\textbf{(b)} AquaMend compares re-probing with admissible terminal recovery decisions under one expected-loss objective on a probe--belief--action graph.
The closed-form threshold applies in the structured regime (Proposition~\ref{prop:threeway}); the general policy is myopic, without a global optimality guarantee.
\textbf{(c)} Earlier discrete T4-Stress results on 4,600 paired episodes per method.
Two-sided paired Wilcoxon comparisons yield Holm-adjusted $p<10^{-3}$ versus restart, decoupled, and greedy, but $p=0.317$ versus depth-2 lookahead (Table~\ref{tab:stress}).
}
\label{fig:overview}
\end{figure}

\section{Introduction}
\label{sec:intro}

\paragraph{Background.}
Embodied agents use \emph{active perception} to acquire latent physical properties, such as cup fill levels, that support planning and execution~\citep{bajcsy1988active,bohg2017interactive}.
These measurements enter a world model as \emph{latent beliefs}, as in belief-space planning and partially observable control~\citep{kaelbling2013integrated,kaelbling1998planning,garrett2021integrated}.
Their validity can expire: covertly adding water changes a cup's fill level, while sensing errors can corrupt evidence without changing the world.
A belief failure may therefore surface as an action error, even when execution follows the stored belief correctly.
Undoing or repeating actions alone does not establish whether the supporting belief remains valid.

\paragraph{Recent work and motivation.}
\emph{Task restart} repeats execution and may be infeasible if it requires undoing an irreversible action.
\emph{Dependency-graph rollback} undoes causally dependent operations~\citep{elnozahy2002survey,haerder1983principles}; truth-maintenance systems retract unsupported conclusions~\citep{doyle1979truth,dekleer1986assumption}, while execution monitoring and plan repair revise plans and states~\citep{fikes1972learning,fox2006plan}.
Decision-theoretic troubleshooting already weighs observations against repairs~\citep{breese1996decision}; our question connects that trade-off to physical belief support and executed-action dependencies.
\textbf{(A) The recovery target is a belief.}
A physically acquired latent property can be checked before selecting downstream correction targets, although exact source identification is not required for feasible rollback.
\textbf{(B) Re-probing complements rollback.}
A measurement can restore belief support or reveal a need for correction, trading sensing cost against potentially avoidable rollback.
Rollback-only baselines lack this information-gathering option, whereas DTT already includes observation decisions.
\textbf{(C) Linear-chain degeneracy is mitigated.}
In a linear causal chain, the minimal closure of an early belief contains the entire executed suffix.
AquaMend retains this rollback rule but checks suspected beliefs before choosing correction targets.
Its benefit under a linear chain therefore comes mainly from avoiding rollback on false alarms, rather than shrinking a required rollback set.

\paragraph{Our work.}
We propose AquaMend, which places physical probes, latent beliefs, and actions in one recovery dependency graph.
A joint posterior guides a one-step comparison of re-probing with admissible terminal recovery decisions, including supported continuation and feasible rollback.
The objective accounts for sensing, executed physical recovery, and genuine failures left uncorrected.
Conditional detection-power screening limits candidate evaluation where its coverage conditions hold; otherwise, all legal probes are retained.

\paragraph{Contributions.}
\begin{itemize}
\item \textbf{Formalization.}
A probe--belief--action graph represents physical re-support of beliefs and evaluates information acquisition and recovery under one expected-loss objective (Sections~\ref{sec:prelim}--\ref{sec:model}).
\item \textbf{Closed-form optimality in a structured regime.}
Under the stated independence, sensing, and separability assumptions, optimal decisions decouple per belief into keep, rollback, or re-probe (Proposition~\ref{prop:threeway}).
\item \textbf{A myopic policy for the general case.}
A joint-posterior policy handles correlated failures, shared recovery costs, and partially resolving probes, without a global optimality or approximation-ratio guarantee (Section~\ref{sec:general}).
\item \textbf{Evaluation.}
Paired pouring simulations assess recovery, re-probing, causal granularity, and computation; separate small-graph and posterior-sensitivity studies examine planning and probability-model limitations (Section~\ref{sec:experiments}).
\end{itemize}

\section{Preliminaries}
\label{sec:prelim}

\paragraph{A dependency graph with three node types.}
The directed acyclic recovery graph $G=(V,E)$ partitions $V=\Mset\cup\Phi\cup\Aset$ into probes, latent beliefs, and actions (Table~\ref{tab:node_types_main}).
Edges encode measurement support $m\to\varphi$, derived support $\varphi_j\to\varphi_i$, belief-to-action dependence $\varphi\to a$, and action dependencies $a\to a'$.
A probe can re-support a belief instead of requiring downstream rollback; edge-type names are specified in Appendix~\ref{app:details}.

\begin{table}[t]
\caption{Node types of the recovery dependency graph.}
\label{tab:node_types_main}
\begin{center}
\begin{tabular}{llll}
\toprule
\textbf{Type} & \textbf{Symbol} & \textbf{Semantics} & \textbf{Cost} \\
\midrule
Probe  & $m\in\Mset$   & one latent-property measurement & redo $\cre(m)\ge0$ \\
Belief & $\varphi\in\Phi$ & latent proposition (e.g., ``cup 3 is least full'') & residual $L_i\ge0$ \\
Action & $a\in\Aset$   & e.g., ``pour into cup 3'' & rollback $\crb(a)\in[0,\infty]$ \\
\bottomrule
\end{tabular}
\end{center}
\end{table}

\paragraph{Downstream rollback and re-probe costs.}
For executed actions $\Aset_{\mathrm{exec}}\subseteq\Aset$, define belief $\varphi_i$'s downstream set and rollback cost as
\begin{equation}
D_i:=\{a\in\Aset_{\mathrm{exec}}:\ \exists\ \text{directed path}\ \varphi_i\rightsquigarrow a\},
\qquad B_i:=C_R(D_i)\in[0,+\infty].
\end{equation}
Here $C_R(R)$ charges physical recovery operations actually performed, including rollback, re-grasping, and re-execution.
Shared operations and overlapping parent--child intervals are charged once; actual re-executions incur new costs.
Under additive accounting, these charges sum over distinct operations at nonnegative cost.
Unexecuted actions are excluded from $D_i$ and may be cancelled at zero cost.
An executed irreversible action in $D_i$ makes that rollback infeasible ($B_i=\infty$).
A re-probe may rule out a false alarm; correction requiring irreversible undo must escalate.
The re-probe cost is $r_i:=\sum_{m\in M_i}\cre(m)$ for a supplied set $M_i$ sufficient to re-support $\varphi_i$, or the expected resolution cost of a sequential prober.

\paragraph{Task-relevant belief failure.}
A belief requires correction if it exceeds task error tolerances or its continued use can alter a downstream decision.
Define
\[
Z_i=\mathbf{1}\{\varphi_i\text{ requires correction for the current task}\},
\qquad \Phi^\star=\{\varphi_i:Z_i=1\}.
\]
Sensor error can yield $Z_i=1$ without physical change, whereas a false alarm need not indicate task-relevant failure.
At information state $x$, $\hat q_i(x)=\sum_s p(s\mid x)Z_i(s)$ is the model's posterior failure probability.
For declaration version $b^v$, we separately query $q_b^v(x)=\sum_s p(s\mid x)\mathbf{1}\{b^v\text{ is invalid in }s\}$.
These proposition-specific probabilities neither measure task success nor guarantee usable supporting evidence.

\paragraph{Failure trigger, attribution, and posterior inference.}
A monitor flags $\varphi^-$ as suspected-invalid or stale, with attribution categories
\begin{equation}
\attr_i\in\{\selferr,\ \external,\ \unknown,\ \falsealarm\}.
\end{equation}
\selferr{} denotes mis-estimation or biased sensing and motivates cross-modal re-probing when an admissible alternative exists.
\external{} denotes physical change; unresolved \unknown{} attribution is handled conservatively, while \falsealarm{} is chiefly an evaluation label.
Ground-truth attribution is not supplied to the online policy; continuation requires available belief support and compliance with task constraints.
The execution-level implementation uses a frozen finite joint model of physical changes and persistent sensing errors.
Public evidence conditions its posterior, incorporating the common alarm once and subsequent observations through Bayesian updates.
Post-hoc Brier scores and expected calibration error assess probability quality without fitting a score-to-probability calibrator (Appendix~\ref{app:q1}).
Structured optimality remains conditional on this supplied probability model and the stated assumptions.

\paragraph{Recovery objective.}
Let $\Phi^-$ contain suspected beliefs and $L_i\ge0$ penalize genuine failure $Z_i=1$ left uncorrected; feasibility restrictions are enforced separately.
An adaptive policy $\pi$ maps observation histories to re-probing, recovery, or stopping decisions to minimize
\begin{equation}
\label{eq:objective}
\min_{\pi}\ \mathbb{E}\!\left[
\underbrace{\sum_{m\in M_\pi}\cre(m)}_{\text{sensing cost}}
+\underbrace{C_{\mathrm{phys}}(\pi)}_{\text{physical cost}}
+\underbrace{\sum_{i:\ Z_i=1,\ \text{uncorrected}} L_i}_{\text{residual penalty}}
\right].
\end{equation}
Here $M_\pi$ records executed sensing events, including repetitions, with total cost $C_P(\pi)=\sum_{m\in M_\pi}\cre(m)$.
The physical cost $C_{\mathrm{phys}}(\pi)=C_{\mathrm{rb}}(\pi)+C_{\mathrm{cont}}(\pi)$ covers rollback and continuation, including operations in $C_R(R)$ without charging them again.
AquaMend selects the minimal upper closure for its correction targets; feasible larger rollback sets remain admissible at their ordinary cost, without an extra mistaken-rollback penalty.
Rollback requires physical feasibility and reversibility, but not prior identification of the exact failure source.

\section{The AquaMend Model}
\label{sec:model}

\subsection{Joint recovery algorithm}
\label{sec:algo}
Algorithm~\ref{alg:jointrecover} uses the joint posterior to select re-probes by one-step expected loss reduction.
Let $V_{\mathrm{stop}}(x)$ denote the minimum expected remaining loss when information acquisition stops and the best admissible terminal decision is taken at state $x$.
A candidate re-probe $m$ has expected net benefit $V_{\mathrm{stop}}(x)-\bigl(\cre(m)+\mathbb{E}[V_{\mathrm{stop}}(x')\mid x,m]\bigr)$.
The policy executes the evaluated probe with the largest positive benefit; otherwise, it commits to the best admissible terminal decision.
The closed-form structured analysis applies only under Assumptions~\ref{ass:a1}--\ref{ass:a3} and finite rollback costs (Section~\ref{sec:prop1}).

\begin{algorithm}[t]
\caption{\textsc{JointRecover}$(G,\text{triggers})$}
\label{alg:jointrecover}
\begin{algorithmic}[1]
\State $I\gets\text{triggers}\subseteq\Phi$
\State $x\gets\textsc{InitJointState}(G,\text{public evidence})$
\State $x\gets\textsc{ConditionOnAlarmOnce}(x,\text{triggers})$
\For{each $\varphi_i\in I$}
  \State $\hat q_i\gets\Pr(Z_i=1\mid x)$
  \State $(r_i,B_i,L_i)\gets\textsc{LocalCosts}(x,\varphi_i)$
\EndFor
\If{structured branch enabled, Assumptions~\ref{ass:a1}--\ref{ass:a3} hold, and all $B_i<\infty$}
  \Comment{optional structured case}
  \For{each $\varphi_i\in I$}
    \State $C_i^{\mathrm{keep}}\gets+\infty$; $\bar L_i\gets+\infty$
    \If{$\textsc{KeepAdmissible}(x,\varphi_i)$}
      \State $C_i^{\mathrm{keep}}\gets\hat q_iL_i$; $\bar L_i\gets L_i$
    \EndIf
    \State $C\gets\{\text{keep}:C_i^{\mathrm{keep}},\ \text{rollback}:B_i,\ \text{reprobe}:r_i+\hat q_i\min(B_i,\bar L_i)\}$
    \State $d_i^\star\gets\arg\min C$
    \State \textsc{ExecuteLocalDecision}$(d_i^\star)$
      \Comment{after probing, act on the outcome}
  \EndFor
  \State \textbf{return}
\EndIf
\While{\textbf{true}}
  \State Refresh belief declarations and available support at $x$
  \State $u^\star\gets\arg\min_{u\in\mathcal T(x)}\sum_s p(s\mid x)\ell(u,s)$
  \State $V\gets V_{\mathrm{stop}}(x)$
  \State $C\gets\textsc{LegalCandidateProbes}(x)$
  \State $C'\gets C$
  \If{$\textsc{ScreeningCovered}(C,x)$}
    \State $C'\gets\textsc{Screen}(C,x)$
  \EndIf
  \If{$C'=\varnothing$}
    \State \textsc{Commit}$(u^\star)$; \textbf{return}
  \EndIf
  \For{$m\in C'$}
    \State $g(m)\gets V-\bigl(\cre(m)+\mathbb{E}[V_{\mathrm{stop}}(x')\mid x,m]\bigr)$
  \EndFor
  \State $m^\star\gets\arg\max_{m\in C'}g(m)$
  \If{$g(m^\star)\leq0$}
    \State \textsc{Commit}$(u^\star)$; \textbf{return}
  \Else
    \State $y\gets\textsc{ExecuteReprobe}(m^\star)$
    \State $x\gets\textsc{BayesUpdate}(x,m^\star,y)$
  \EndIf
\EndWhile
\end{algorithmic}
\end{algorithm}

\subsection{Structured regime: a per-belief three-way threshold}
\label{sec:prop1}
The structured regime decouples under three assumptions.

\begin{assumption}[Independent, disjoint downstreams]
\label{ass:a1}
Failure indicators $Z_i$ are mutually independent, and downstream sets $D_i$ are pairwise disjoint.
Admissible probe sets are belief-local and disjoint: conditional on $Z_i$, probing belief $i$ provides no information about other beliefs.
Probe costs and terminal decisions are separable, with no binding shared recovery budget.
\end{assumption}
\begin{assumption}[Additive DAG rollback costs]
\label{ass:a2}
The causal subgraph is a DAG, with nonnegative recovery costs additive over distinct charged operations.
Shared operations and overlapping parent--child intervals cannot contribute duplicate charges, and distinct beliefs share no charged recovery operations.
Feasible corrective recovery removes the local residual failure and requires closure $D_i$, costing $B_i$ after subtracting continuation costs common to the choices.
Larger feasible bundles remain admissible at their ordinary nonnegative additional cost.
\end{assumption}
\begin{assumption}[Fully resolving re-probe]
\label{ass:a3}
The designated probe for $\varphi_i$ fully resolves $Z_i$ at cost $r_i$, using a supplied admissible probe set.
It changes neither the failure state nor terminal costs; a resolved valid belief permits continuation with zero additional local correction cost.
\end{assumption}

\begin{proposition}[Three-way threshold optimality]
\label{prop:threeway}
Under Assumptions~\ref{ass:a1}--\ref{ass:a3} and finite $B_i$, the expected loss in Eq.~\eqref{eq:objective} is separable across beliefs.
The following local costs omit continuation terms common to the compared choices.
Define $\bar L_i=L_i$ if keep-and-ignore is admissible and $\bar L_i=+\infty$ otherwise.
The optimal local decision minimizes
\begin{equation}
\label{eq:threeway}
C_i^{\mathrm{keep}},\qquad B_i,\qquad
r_i+\hat q_i\min\{B_i,\bar L_i\},
\end{equation}
where $C_i^{\mathrm{keep}}=\hat q_iL_i$ when keep-and-ignore is admissible and $+\infty$ otherwise.
If $B_i\leq L_i$ or keep-and-ignore is inadmissible, the probe branch costs $r_i+\hat q_iB_i$, giving
\begin{equation}
\text{re-probe}\ \succ\ \text{direct rollback}\iff r_i<(1-\hat q_i)B_i.
\end{equation}
Comparisons with admissible keep-and-ignore follow directly from Eq.~\eqref{eq:threeway} (Appendix~\ref{app:prop1}).
Equality denotes a tie, not a strict preference.
Conditional on choosing corrective rollback, the minimal upper closure is used.
With local costs supplied, the decisions require $O(|\Phi^-|)$ scalar comparisons; graph preprocessing is discussed in Appendix~\ref{app:prop1}.
If correction requires undoing an already-executed irreversible action, that rollback is infeasible and recovery must escalate.
\end{proposition}

Appendix~\ref{app:prop1} proves Proposition~\ref{prop:threeway}.
Intuitively, re-probing pays $r_i$ for probability $1-\hat q_i$ of avoiding rollback $B_i$, yielding the threshold $r_i<(1-\hat q_i)B_i$.

\begin{remark}[Scope of Proposition~\ref{prop:threeway}]
\label{rem:a1}
T4 violates Assumption~\ref{ass:a1}: adding water or swapping cups correlates ranking beliefs, and pouring depends on their joint ranking through shared downstream actions.
The proposition therefore does not establish optimality for the general policy evaluated on T4.
\end{remark}

\subsection{General regime: a myopic adaptive policy}
\label{sec:general}
Correlated failures, shared recovery operations, and partially resolving probes require joint decisions.
At information state $x$, containing evidence and recovery history, define
\begin{equation}
\label{eq:terminalvalue}
V_{\mathrm{stop}}(x)=\min_{u\in\mathcal T(x)}\sum_s p(s\mid x)\ell(u,s),
\end{equation}
where $p(s\mid x)$ covers joint physical changes and persistent sensing errors, and $\mathcal T(x)$ contains admissible terminal decisions.
The loss $\ell(u,s)$ includes remaining sensing, recovery, continuation, and residual failure costs; incurred costs are sunk.
Each candidate decision $u$ must apply across all states still possible at $x$.
Algorithm~\ref{alg:jointrecover} evaluates one probe followed by a terminal decision, without optimizing a complete adaptive policy tree.

\paragraph{Computing the greedy step.}
The execution-level implementation enumerates finite joint configurations and predictive observation branches, updating the posterior and terminal decision for each outcome.
Legacy benchmarks also evaluate residual rollback risk
\[
U(x)=\sum_{S\subseteq\Phi}\Pr(S\mid x)B(S),
\]
where $B(S)$ is the minimal closure cost for failure set $S$, charging shared operations once.
This evaluates a fixed closure-rollback rule and generally differs from the optimized terminal value $V_{\mathrm{stop}}(x)$.
For sensing that leaves $S$ and $B(S)$ unchanged, $\mathbb{E}[U(x')\mid x,m]=U(x)$; expected reduction in $U$ alone cannot represent task-loss value of information.
Legacy M1 uses exact evaluation for $|V|\leq12$ and $|\Phi|\leq5$; rollback-risk approximations appear in Table~\ref{tab:utility-proxies} and Appendix~\ref{app:details}.

\begin{table}[t]
\caption{Computable proxies for residual rollback risk $U(x)$ in the earlier evaluations.
Evaluation complexities assume supplied marginal probabilities and local costs, retained hypotheses and weights, or posterior samples.
$T_B$ denotes the time required to evaluate one closure cost $B(S)$.}
\label{tab:utility-proxies}
\begin{center}
\resizebox{\linewidth}{!}{%
\begin{tabular}{lll}
\toprule
\textbf{Mode} & \textbf{Definition} & \textbf{Complexity} \\
\midrule
Indep
& $\tilde U=\sum_i\hat q_iB_i$ for disjoint charged recovery operations
& $O(|\Phi|)$ \\
Beam-$K$
& approximate $U(x)$ using up to $K$ weighted joint failure hypotheses
& $O(KT_B)$ \\
MC
& average $B(S)$ over $N_{\mathrm{mc}}$ posterior samples of the failure set
& $O(N_{\mathrm{mc}}T_B)$ \\
\bottomrule
\end{tabular}%
}
\end{center}
\end{table}

We claim neither global optimality nor an approximation-ratio guarantee for the general-regime policy.

\subsection{Re-probing as two substeps}
\label{sec:reprobe}
\textbf{(i) Detection} tests the fixed pre-probe declaration $b^v$ using model-conditional power
\begin{equation}
C_s(m,b^v\mid x)=\Pr\!\left(q_b^v(x,m,Y)\geq0.5\mid b^v\text{ invalid},x,m\right),
\end{equation}
where $Y$ is the future observation and the conditioning event has positive probability.
When all legal candidates satisfy screening coverage, screening retains per-declaration power-per-cost maximizers plus candidates required for re-estimation or evidence-source replacement; otherwise, it retains all candidates.
Algorithm~\ref{alg:jointrecover} compares retained probes with admissible terminal decisions under expected loss.
\textbf{(ii) Re-estimation} updates the property and checks support for the refreshed declaration, a distinct query from testing the old version.
Coverage conditions and implementation details are given in Appendix~\ref{app:screening}.

\section{Experiments}
\label{sec:experiments}
\subsection{Platforms and data}
\textbf{LatentProp-Bench task T4} is a time-stepped multi-cup pouring simulation.
Episodes acquire evidence, execute an action prefix, encounter a controlled perturbation, and trigger recovery before continued execution or stopping.
Separate earlier discrete T4 and M1 evaluations retain their original protocols.

\subsection{Perturbations and causal granularity}
The execution-level evaluation crosses four perturbation families (Table~\ref{tab:perturb}) with early and late stages, using four independent scenario seeds per combination.
These 32 scenarios are paired across methods; fine-causal and linear-chain controls distinguish avoided rollback from smaller rollback sets (Table~\ref{tab:gran}).

\begin{table}[t]
\caption{Perturbation families and ground-truth attribution.}
\label{tab:perturb}
\begin{center}
\normalsize
\setlength{\tabcolsep}{4pt}
\begin{tabularx}{\linewidth}{
  @{}l
  >{\raggedright\arraybackslash}X
  >{\raggedright\arraybackslash}X
  l@{}
}
\toprule
\textbf{Family} & \textbf{Operation} & \textbf{Failed beliefs $\Phi^\star$} & \textbf{Attribution} \\
\midrule
Swap cups     & swap two containers' poses/IDs & water level / target of swapped instances & \external \\
Add water     & add $\Delta V$ across a decision boundary & filled cup's level and ranking & \external \\
Sensor drift  & bias / timestamp / mis-binding (world unchanged) & beliefs on the corrupted evidence & \selferr \\
False alarm   & brief occlusion / noise spike (no task-relevant failure) & $\emptyset$ & \falsealarm \\
\bottomrule
\end{tabularx}
\end{center}
\end{table}

\begin{table}[!htbp]
\caption{Two causal-edge granularities and their expected effects.}
\label{tab:gran}
\begin{center}
\normalsize
\setlength{\tabcolsep}{4pt}
\begin{tabularx}{\linewidth}{
  @{}l
  >{\raggedright\arraybackslash}X
  >{\raggedright\arraybackslash}X
  @{}
}
\toprule
\textbf{Granularity} & \textbf{Edge semantics} & \textbf{Expectation} \\
\midrule
Linear chain & stage dependency; rollback = suffix & benefit mainly from zero rollback on false alarms \\
Fine causal  & explicit causal/evidence edges; rollback = minimal upper closure & rollback set itself is smaller \\
\bottomrule
\end{tabularx}
\end{center}
\end{table}

\subsection{Metrics}
Complete success ($\mathrm{Succ}$) requires physical completion, valid belief support, and compliance with quality and budget constraints; safe stopping alone is unsuccessful.
Primary complete loss is $\mathcal L_n=C_n+\sum_{k\in\Phi_n^{\mathrm{res}}}L_k$, with $C_n=C_{P,n}+C_{\mathrm{rb},n}+C_{\mathrm{cont},n}$ and $\Phi_n^{\mathrm{res}}$ containing genuine uncorrected failures.
Operational saving $\mathrm{Save}_C=1-\sum_n C_n/\sum_n C_{n,\mathrm{Restart}}$ excludes residual losses.
We also report re-probe and rollback counts/costs ($\#P,C_P$; $\#R,C_{\mathrm{rb}}$), online decision time, and false-alarm zero-rollback rate (ZRR), including its conjunction with success.
Physical costs use simulation seconds, complete loss uses simulation-second equivalents, and online decision time uses wall-clock seconds.

\subsection{Baselines, ablations, and protocol}
Legacy comparisons cover restart/retry, rollback, troubleshooting~\citep{breese1996decision}, belief-space replan, TMS revision~\citep{doyle1979truth}, decoupled/greedy/lookahead policies, and an oracle (Tables~\ref{tab:main}, \ref{tab:stress}, and~\ref{tab:m1}).
M1 additionally uses Exact Adaptive, which solves the complete finite policy tree from the public posterior.
Only the oracle accesses $\Phi^\star$; its lower-bound interpretation requires the conditions in Appendix~\ref{app:prop2}.
Earlier ablations vary re-probing, attribution, source separation, causal granularity, probe scoring, noise, posteriors, and residual penalties (Appendix~\ref{app:legacy}).

The execution-level comparison includes AquaMend, full restart, DTT, no extra re-probe, and a linear-chain control, yielding 160 executions across the 32 scenarios.
They share action primitives, physical accounting, safety constraints, and success criteria.
A separate all-candidate ablation (no-$C_s$/netVoI-all) uses AquaMend's posterior and one-step objective without screening.

Two-sided paired Wilcoxon signed-rank tests assess losses, with Holm correction against AquaMend within each designated comparison family.
Standard T4 uses its reported family; T4-Stress separates four designated non-oracle comparisons from five supplemental traditional-baseline comparisons.
The execution-level family contains the four non-AquaMend methods listed above.
Unadjusted 95\% intervals for mean paired loss differences use 2,000 ordinary paired bootstrap resamples of the 32 scenarios and percentile endpoints.
All successful, failed, and partial executions are retained; irreversible-action cases are reported separately, and incomplete escalations count as failures.
Discrete and execution-level protocols, costs, and sample counts remain separate (Appendices~\ref{app:legacy} and~\ref{app:newprotocol}).

\paragraph{Discrete recovery on LatentProp-Bench task T4.}
The 4,900-episode reversible standard-T4 split did not distinguish AquaMend, decoupled, greedy, or depth-2 policies.
T4-Stress adds common-cause failures, shared closures, and sensing budgets across 4,600 paired episodes per method (Figure~\ref{fig:overview}c).
AquaMend achieved success 0.976 and mean loss 12.47, with significantly lower paired losses than decoupled recovery, greedy re-probing, and restart after Holm correction.
The internal depth-2 planning control matched AquaMend's
rounded success and mean loss, with no statistically
significant paired loss difference ($p_{\mathrm{Holm}}=0.317$).
This comparison does not establish equivalence between
the policies or rule out benefits from deeper planning
on other tasks.
Separate hard rank-tie, adversarial-conflict, and shared-closure diagnostics yielded success 0.025, 0.425, and 0.780; they are excluded from the primary average (Appendix~\ref{app:legacy}).

\begin{figure}[t]
\centering
\includegraphics[width=\linewidth]{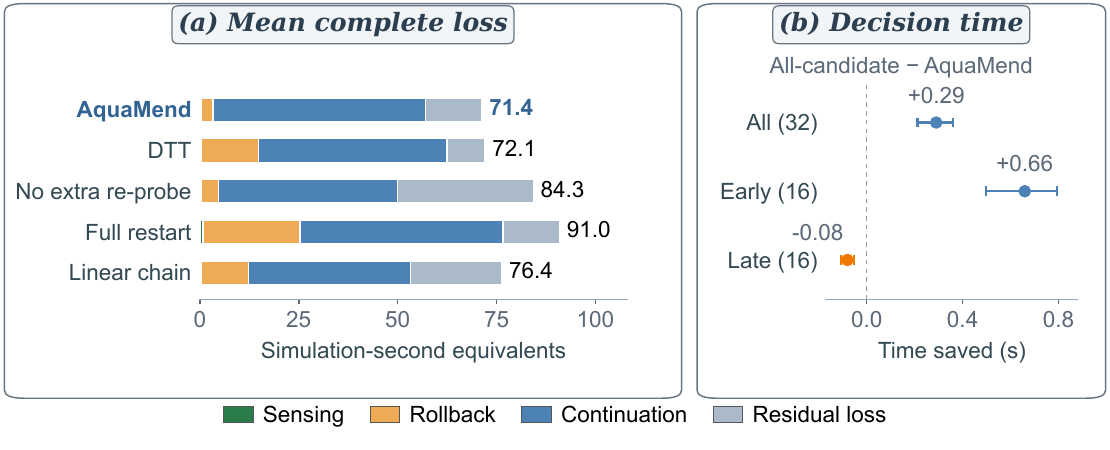}
\par\vspace{-2pt}
\caption{\textbf{Execution-level loss and decision time.}
\textbf{(a)} Mean loss components for 32 paired scenarios per method; labels show reported totals, subject to rounding.
\textbf{(b)} Mean paired decision-time differences (all-candidate minus AquaMend), with unadjusted 95\% confidence intervals; positive values favor AquaMend.
Each scenario contributes the median of four complete calls ($n=32$ overall; $n=16$ per stage).
Source values appear in Tables~\ref{tab:cost160} and~\ref{tab:timing160}.}
\label{fig:physical_tradeoff}
\end{figure}

\paragraph{Execution-level recovery.}
In the separate 32-scenario evaluation, AquaMend recovered in 28 cases versus DTT's 25 and reduced mean complete loss by 21.6\% relative to restart.
The paired analysis supports lower loss than restart but does not establish an advantage over DTT after Holm correction (Table~\ref{tab:physical160}).
Operational saving was 25.6\%, excluding residual losses; Figure~\ref{fig:physical_tradeoff}a decomposes complete loss.

\begin{table}[t]
\caption{Recovery in the execution-level simulation, with 32 paired scenarios per implementation.
Complete loss is measured in simulation-second equivalents.
$\Delta\mathcal{L}$ is comparator loss minus AquaMend loss, with unadjusted 95\% paired-bootstrap intervals; Holm correction covers the four comparisons.}
\label{tab:physical160}
\begin{center}
\setlength{\tabcolsep}{4pt}
\begin{tabular*}{\linewidth}{@{\extracolsep{\fill}}lcccc@{}}
\toprule
\textbf{Method} & $\mathrm{Succ}\uparrow$ & $\overline{\mathcal{L}}\downarrow$ & $\Delta\mathcal{L}$ [95\% CI] & $p$ (Holm) \\
\midrule
\textbf{AquaMend (ours)} & \textbf{28/32} & \textbf{71.356} & -- & -- \\
DTT & $25/32$ & $72.116$ & $0.761\;[-7.571,\,7.901]$ & $0.465$ \\
No extra re-probe & $21/32$ & $84.326$ & $12.970\;[1.014,\,27.115]$ & $0.233$ \\
Full restart & $24/32$ & $90.966$ & $19.610\;[12.944,\,26.523]$ & $8.09\times10^{-5}$ \\
Linear chain & $22/32$ & $76.391$ & $5.035\;[-0.568,\,10.018]$ & $0.077$ \\
\bottomrule
\end{tabular*}
\end{center}
\end{table}

\paragraph{Ablations: re-probing and causal granularity.}
Removing re-probing increased mean loss from 5.431 to 38.157
and reduced success from 0.901 to 0.213 in the earlier discrete
T4 ablation (Figure~\ref{fig:reprobe_ablation});
causal-granularity results remain in
Appendix~\ref{app:legacy}, Table~\ref{tab:ablation}.

In the separate 32-scenario execution-level evaluation, removing additional re-probing increased mean complete loss from 71.356 to 84.326, while replacing fine causal dependencies with a linear chain increased it to 76.391 (Table~\ref{tab:physical160}).
Neither comparison met the Holm-adjusted significance threshold.
AquaMend, DTT, the all-candidate ablation, no extra re-probe, and the linear-chain control achieved zero rollback and complete recovery on all eight false-alarm scenarios.
Zero rollback on these false alarms therefore does not distinguish AquaMend from these controls.

\paragraph{Internal planning comparison on M1.}
Exact Adaptive and depth-2 lookahead are internal planning controls sharing AquaMend's finite model, graph, costs, and public posterior.
Across 4,000 graphs and five seeds, AquaMend's mean expected-risk gap to Exact Adaptive was 0.037 (Table~\ref{tab:m1}).
This does not establish a general approximation or optimality guarantee.
In the six-candidate M1-Depth diagnostic, AquaMend and depth-2/depth-3 policies incurred risk 4.000 versus 3.750 for Exact Adaptive (Appendix~\ref{app:m1depth}).

\begin{figure}[t]
\centering
\includegraphics[width=\linewidth]{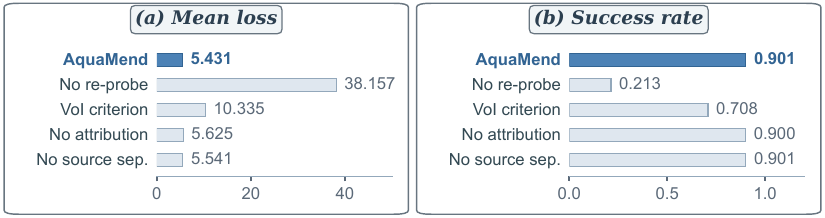}
\par\vspace{-2pt}
\caption{\textbf{Component ablations on discrete T4.}
Internal ablation controls on the earlier 6,100-episode set, including its irreversible subset, under the legacy accounting protocol (Table~\ref{tab:ablation}).
VoI denotes value of information; ``No source sep.'' merges self and external sources.}
\label{fig:reprobe_ablation}
\end{figure}

\paragraph{Online decision time.}
AquaMend and the all-candidate ablation matched physical outcomes in the original batch.
Under identical caching, screening reduced mean decision time by 12.3\% overall but increased it by 3.4\% late, where all candidates were evaluated (Figure~\ref{fig:physical_tradeoff}b; Appendix~\ref{app:newprotocol}).

\paragraph{Posterior sensitivity.}
A separate eight-scenario study retained recovery in 7/8 cases under all five posterior settings, although dispersion and positive failure bias increased mean loss.
These 40 executions provide a limited sensitivity check, not general robustness evidence, and are not pooled with the main evaluation (Appendix~\ref{app:q1}).

\section{Related Work}
\label{sec:related}

\paragraph{Diagnosis and rollback.}
Troubleshooting weighs observation against repair~\citep{heckerman1995troubleshooting,breese1996decision}, while test sequencing orders diagnostic observations~\citep{pattipati1990test}.
Decision-tree and adaptive-coverage results~\citep{hyafil1976constructing,golovin2011adaptive,guillory2010interactive} do not directly establish guarantees for our recovery objective.
Dependency-based rollback~\citep{elnozahy2002survey,haerder1983principles} and truth maintenance~\citep{doyle1979truth,dekleer1986assumption} provide foundations for recovering executed actions and reconsidering unsupported beliefs.

\paragraph{Planning, perception, and robot recovery.}
Belief-space planning handles uncertainty~\citep{kaelbling1998planning,kaelbling2013integrated,garrett2021integrated}; active perception acquires information through action~\citep{bajcsy1988active,bohg2017interactive}; execution monitoring and plan repair address execution deviations~\citep{fikes1972learning,fox2006plan}.
REFLECT explains failures to guide corrective planning~\citep{liu2023reflect}, AIC MLLM corrects manipulation poses using interaction feedback~\citep{xiong2025aic}, and human-assisted recovery balances module uncertainty against intervention costs~\citep{banerjee2026modularhil}.
AquaMend connects troubleshooting's observation--repair trade-off to physical belief support and executed-action dependencies through one expected-loss objective.

\section{Conclusion}
\label{sec:conclusion}

AquaMend couples physical re-probing, belief support, and executed-action rollback under one expected-loss objective.
The closed-form optimum requires structured assumptions; the general policy is myopic, without a global optimality guarantee.
Simulations support lower complete loss than restart, but do not establish a loss advantage over DTT.
Screening saves decision time overall but adds overhead in the uncovered late stage.
Evidence is limited to self-constructed simulations without an external public leaderboard or physical-robot validation.
Future work includes general-regime guarantees, posterior robustness, and physical deployment.

\subsection*{AI use statement}
AI tools assisted with English editing, LaTeX revision, figure layout refinement, and cross-section consistency checks.
They also assisted with literature discovery and identification of relevant work, auditing the scope of theoretical claims,
and interpretation of author-supplied results.
The authors remain responsible for verifying all AI-assisted material and for the manuscript's claims, data, analyses, references, and final content.

\subsection*{Reproducibility statement}

The model, algorithm, and evaluation protocol are described in Sections~\ref{sec:prelim}--\ref{sec:experiments}.
Appendix~\ref{app:prop1} provides the structured-regime proof.
Benchmark details are given in Appendices~\ref{app:details},
\ref{app:m1depth}, and~\ref{app:legacy}; the paired evaluation,
posterior-sensitivity study, and screening implementation are described in Appendices~\ref{app:newprotocol}--\ref{app:screening}.
Exact computational reproduction requires the frozen implementation and underlying execution records, as discussed in Appendix~\ref{app:newprotocol}.

\bibliography{references}
\bibliographystyle{iclr2027_conference}

\appendix
\section{Proof of Proposition~\ref{prop:threeway}}
\label{app:prop1}

Each suspected belief $\varphi_i$ has a supplied re-probe cost $r_i\geq0$, finite corrective rollback cost $B_i$, posterior failure probability $\hat q_i$, and residual penalty $L_i\geq0$.
The proof uses Assumptions~\ref{ass:a1}--\ref{ass:a3}, with continuation costs common to the local choices omitted.

\paragraph{Step 1: Additivity.}
Belief-local probes, independent failure indicators, disjoint charged recovery operations, and separable terminal decisions make the recovery choices local to each belief.
There is no binding shared budget.
The total expected loss therefore decomposes as
\[
\mathbb{E}[C(\pi)]
=
\sum_i \mathbb{E}[C_i(\pi_i)].
\]
Minimizing the total is equivalent to minimizing each local expected loss.
Fix a belief and suppress its index, writing $\hat q,r,B,L$.

\paragraph{Step 2: Single-belief expectations.}
Without probing, keep-and-ignore costs $\hat qL$ when admissible and $+\infty$ otherwise.
Direct corrective rollback costs $B$ and removes the local residual failure.
Define $\bar L=L$ when keep-and-ignore is admissible and $\bar L=+\infty$ otherwise.

A designated fully resolving probe costs $r$ and reveals whether the belief requires correction without changing the failure state or terminal costs.
A valid outcome permits continuation with zero additional local correction cost.
A failure outcome is followed by the cheaper of corrective rollback and admissible acceptance, at cost $\min\{B,\bar L\}$.
The expected probe cost is therefore
\[
r+\hat q\min\{B,\bar L\}.
\]
A repeated fully resolving probe supplies no further information and has nonnegative cost.
Thus, the optimal local decision minimizes the three costs in Eq.~\eqref{eq:threeway}.

\paragraph{Step 3: Threshold algebra.}
If $B\leq L$ or keep-and-ignore is inadmissible, then $\min\{B,\bar L\}=B$.
Consequently,
\[
r+\hat qB<B
\quad\Longleftrightarrow\quad
r<(1-\hat q)B.
\]
The expected saving relative to direct rollback is $(1-\hat q)B-r$.

When keep-and-ignore is admissible, its expected cost $\hat qL$ is compared directly with the other two branches.
In particular, if $B>L$, the probe branch costs $r+\hat qL\geq\hat qL$, so probing cannot strictly improve on keeping.
Equality denotes a tie rather than a strict preference.

\paragraph{Step 4: Optimality of the rollback substep.}
Assumption~\ref{ass:a2} specifies that corrective recovery is feasible, removes the local failure, and requires the executed downstream closure $D_i$.
Any admissible corrective bundle therefore includes the required charged operations.
Because operation costs are nonnegative and additive, including additional operations cannot reduce the corrective cost.
Conditional on choosing corrective rollback, the minimal required closure is therefore optimal in this structured regime.

\paragraph{Step 5: Complexity.}
With local costs and probabilities supplied, each belief requires a constant number of scalar comparisons, giving $O(|\Phi^-|)$ decision work.
A graph traversal for one closure takes $O(|V|+|E|)$ time.
Separate traversals for all suspected beliefs give the straightforward bound $O(|\Phi^-|(|V|+|E|))$, excluding additional cost-model evaluation.
Disjoint downstream action sets alone do not exclude repeated traversal of shared intermediate nodes, so they do not establish a globally linear preprocessing bound.
\hfill$\square$

\begin{corollary}[Value of re-probing]
In the regime of Step~3, the expected saving over direct rollback is $(1-\hat q_i)B_i-r_i$.
For finite $B_i\geq0$, this saving is nonincreasing in $\hat q_i$ and nondecreasing in $B_i$.
Re-probing is strictly preferred to direct rollback when $r_i<(1-\hat q_i)B_i$.
Already-executed irreversible actions lie outside this finite-cost comparison.
A re-probe may rule out a false alarm, while correction that requires undoing an irreversible action must escalate.
Cancelling an unexecuted planned action may instead cost zero.
\end{corollary}

\section{Complexity status and conditional bounds}
\label{app:prop2}

\subsection{Complexity status}
The earlier reduction from optimal decision trees assigns zero cost to hypothesis-specific corrective rollback sets.
It does not exclude a feasible union of these sets that corrects the possible failures without first identifying the true hypothesis.
Under the present formulation, additional feasible rollback is charged at its ordinary physical cost, and residual penalties apply only to genuine failures left uncorrected.
A separate penalty for rolling back unaffected actions would change the problem being studied.
The reduction therefore does not establish NP-hardness of the present formulation.
This proof gap does not establish tractability.

\subsection{Relation to adaptive coverage}
Adaptive submodular set-cover guarantees require specific utility, feasibility, and stopping assumptions \citep{golovin2011adaptive,guillory2010interactive}.
Algorithm~\ref{alg:jointrecover} instead compares the expected loss of one probe followed by a terminal decision with the best terminal decision available now.
It stops when no evaluated candidate has positive expected net benefit.
The cited coverage guarantees therefore do not directly apply, and no approximation ratio is claimed for the general-regime policy.

\subsection{Oracle lower bound and the decoupled upper bound}
A full-information optimal reference can observe the latent state before selecting its recovery decisions.
If it has access to the same physical options as an ordinary policy under the same cost accounting, it can imitate that policy, so its optimal expected loss cannot be larger.

The legacy oracle implements a particular rule: it performs no re-probing and executes the minimal corrective closure for the true failure set.
This specified rule gives a lower bound only when the correction is feasible and required or optimal among the comparable informed recovery options.
For example, when accepting a failure is admissible and $B_i>L_i$, paying $L_i$ is cheaper than unconditional rollback even with perfect information.
The closure oracle is therefore a clairvoyant reference unless these additional conditions hold.

Under nonnegative additive costs, $\sum_i B_i$ upper-bounds the corresponding union-closure rollback cost by counting shared operations repeatedly.
This does not make it an upper bound on an arbitrary policy's complete loss.

\section{Complexity, benchmark generation, and ground-truth protocol}
\label{app:details}
\paragraph{Computable residual-risk proxies.}
The three approximation families differ in how they represent
uncertainty over the set of failed beliefs.
Table~\ref{tab:utility-proxies-details} gives supplementary specifications.
The main-text summary is provided in Table~\ref{tab:utility-proxies}.
Here, $K$ denotes the number of retained joint hypotheses and
$N_{\mathrm{mc}}$ denotes the number of posterior samples.

\begin{table}[t]
\setlength{\belowcaptionskip}{10pt}
\centering
\caption{Approximation families for the residual rollback risk
$U(x)$. The displayed Indep expression applies to
pairwise-disjoint rollback sets.}
\label{tab:utility-proxies-details}
\small
\begin{tabular}{@{}p{0.13\linewidth}p{0.57\linewidth}p{0.20\linewidth}@{}}
\toprule
Mode & Specification & Control parameter \\
\midrule

Indep &
Uses posterior marginal belief-failure probabilities.
For pairwise-disjoint rollback sets,
$\tilde U(x)=\sum_i \hat q_i B_i$. &
Marginal probabilities $\hat q_i$ \\
\addlinespace

Beam-$K$ &
Retains up to $K$ highest-probability joint belief-failure
hypotheses, with candidate expansion along correlation edges,
and approximates rollback risk using the retained hypotheses. &
Beam width $K$ \\
\addlinespace

MC &
Estimates rollback risk from $N_{\mathrm{mc}}$ samples
of the joint belief-failure configuration drawn from
the posterior. &
Sample count $N_{\mathrm{mc}}$ \\

\bottomrule
\end{tabular}
\end{table}

\paragraph{Recovery-graph edge types.}
Measurement support uses \textsc{Detection} and \textsc{Aggregation}; derived belief support uses \textsc{Inference}.
Belief-to-action dependence uses \textsc{Belief\_to\_Action}, while action dependencies use \textsc{Action\_Causal} and \textsc{Temporal}.

\paragraph{Complexity summary.}
Collecting one executed downstream closure takes $O(|V|+|E|)$ graph-traversal time.
With local costs supplied, the structured decisions require $O(|\Phi^-|)$ scalar comparisons.
The general policy additionally evaluates candidate probes, predictive observation branches, posterior updates, and admissible terminal decisions.
Its runtime, and that of bounded-lookahead or exact adaptive planning, therefore depends on these state and observation spaces as well as graph size.

\paragraph{Synthetic small-graph (M1) generation.}
Random DAGs with $|V|\in\{6,8,10,12\}$, out-degree $\mathrm{Unif}\{1,2,3\}$ or Erd\H{o}s--R\'enyi
$p\in\{0.15,0.25\}$; node-type ratio probe/belief/action $\approx0.4/0.4/0.2$; $\cre$ log-normal
clipped to $[c_{\min},c_{\max}]=[0.5,5.0]$; $\crb$ from the same family with
$\mathbb{E}[\crb]/\mathbb{E}[\cre]\in[2,5]$; task-critical beliefs have $L_i>0$, others $0$;
$|\Phi^\star|\sim\mathrm{Unif}\{1,2,3\}$ with an extra $20\%$ false alarms; independent and
correlated modes.

\paragraph{Ground-truth annotation.}
For each \texttt{(scene, perturbation, seed)} we record: (1) the true-changed belief set
$\Phi^\star$ (written by the injector, $10\%$ human-audited); (2) the minimal rollback set $R^\star$
by running minimal upper closure on $\Phi^\star$ for each granularity; (3) the restart cost
$C_{\mathrm{restart}}$ by re-executing the full successful demonstration; (4) the minimal re-probe
set $P^\star$ by enumeration (M1 only; \texttt{null} on T4);
and (5) the attribution label in $\{\selferr,\external,\unknown,\falsealarm\}$.

\section{Controlled planning-depth diagnostic (M1-Depth)}
\label{app:m1depth}

\paragraph{Purpose and construction.}
Standard T4 and T4-Stress did not distinguish AquaMend from depth-2 lookahead. M1-Depth therefore isolates the value of planning horizon on complementary
evidence chains. It is a controlled mechanism experiment built on the M1 information-isolation and exact dynamic-programming interface, rather than a
new T4-Stress split.

Each graph contains $n\in\{3,4,5,6\}$ candidate beliefs, exactly one of which has truly changed. Every belief has an independent probe and an independent
rollback action, forming edges
$\text{probe}_i\rightarrow\text{belief}_i\rightarrow\text{action}_i$. Non-oracle policies observe only the graph, public posterior, and costs; the hidden changed belief is revealed only after execution for scoring. Probe costs are centered at $0.3+1.5/n$ with a $\pm3\%$ seed perturbation, rollback costs
are sampled from $[1.8,2.2]$, the residual loss of an unresolved change is $4.0$, and restart costs $99.0$. These costs make an isolated early probe
unattractive while allowing a sufficiently long sequence of negative observations to become beneficial. In the worst case, identifying the changed
belief among $n$ candidates requires $n-1$ probes.

\paragraph{Policies and protocol.}
We compared four policies. \emph{AquaMend Myopic} plans one step and repeats
the \textsc{General} three-action decision after each observation.
\emph{Lookahead (depth 2)} and \emph{Lookahead (depth 3)} expand two and three future observation levels, respectively, and replan with the same horizon after every real observation. Thus, depth limits the current planning horizon rather than the total number of probes in an episode. \emph{Exact Adaptive} solves the
complete finite policy tree and serves only as an exact small-graph reference, not as a variant of AquaMend.

Experiments used seeds $\{0,1,2,3,4\}$ and belief counts
$\{3,4,5,6\}$. All hidden changed-belief realizations under each public posterior were enumerated exactly, producing
$5\times(3+4+5+6)\times4=360$ policy evaluations. We report posterior-weighted success, realized cost, expected policy risk, median decision time, and median
peak Python memory allocation.

\begin{table}[t]
\setlength{\belowcaptionskip}{10pt}
\caption{Controlled planning-depth results on M1-Depth. Worst-case depth is the number of probes required to identify the final candidate. Decision time
and peak memory report medians over deterministic policy evaluations.}
\label{tab:m1depth}
\begin{center}
\normalsize
\setlength{\tabcolsep}{3pt}
\begin{tabularx}{\linewidth}{
  @{}cc
  >{\raggedright\arraybackslash}X
  rrrrr@{}
}
\textbf{Beliefs} &
\shortstack{\textbf{Worst-case}\\\textbf{depth}} &
\textbf{Policy} &
$\mathrm{Succ}\uparrow$ &
\shortstack{\textbf{Realized}\\\textbf{cost}$\downarrow$} &
\shortstack{\textbf{Expected}\\\textbf{risk}$\downarrow$} &
\shortstack{\textbf{Time}\\\textbf{(ms)}$\downarrow$} &
\shortstack{\textbf{Memory}\\\textbf{(KiB)}$\downarrow$} \\
\midrule
3 & 2 & AquaMend Myopic & 0.800 & 3.396 & 3.396 & 0.512 & 22.4 \\
3 & 2 & Lookahead (depth 2)    & 1.000 & 3.248 & 3.248 & 0.705 & 22.6 \\
3 & 2 & Lookahead (depth 3)    & 1.000 & 3.248 & 3.248 & 0.744 & 39.3 \\
3 & 2 & Exact Adaptive  & 1.000 & 3.248 & 3.248 & 0.533 & 19.0 \\
\midrule
4 & 3 & AquaMend Myopic & 0.000 & 4.000 & 4.000 & 1.013 & 28.2 \\
4 & 3 & Lookahead (depth 2)    & 0.600 & 3.679 & 3.679 & 2.931 & 54.9 \\
4 & 3 & Lookahead (depth 3)    & 1.000 & 3.484 & 3.484 & 3.569 & 78.3 \\
4 & 3 & Exact Adaptive  & 1.000 & 3.484 & 3.484 & 2.455 & 43.4 \\
\midrule
5 & 4 & AquaMend Myopic & 0.000 & 4.000 & 4.000 & 3.583 & 104.3 \\
5 & 4 & Lookahead (depth 2)    & 0.000 & 4.000 & 4.000 & 8.136 & 79.3 \\
5 & 4 & Lookahead (depth 3)    & 0.400 & 3.815 & 3.815 & 10.646 & 358.9 \\
5 & 4 & Exact Adaptive  & 1.000 & 3.594 & 3.594 & 11.841 & 239.8 \\
\midrule
6 & 5 & AquaMend Myopic & 0.000 & 4.000 & 4.000 & 12.577 & 251.0 \\
6 & 5 & Lookahead (depth 2)    & 0.000 & 4.000 & 4.000 & 30.483 & 726.8 \\
6 & 5 & Lookahead (depth 3)    & 0.000 & 4.000 & 4.000 & 44.932 & 1206.8 \\
6 & 5 & Exact Adaptive  & 1.000 & 3.750 & 3.750 & 54.056 & 1487.6 \\
\bottomrule
\end{tabularx}
\end{center}
\end{table}

\paragraph{Results.}
The results followed the expected ordering with respect to planning horizon. For three beliefs, depth-2 lookahead already covered the complete evidence
chain and matched Exact Adaptive at an expected risk of $3.248$. For four beliefs, depth-3 lookahead matched Exact Adaptive at $3.484$, whereas depth-2
lookahead retained a higher risk of $3.679$.

When the complementary evidence chain extended beyond the visible horizon, finite-depth policies stopped and accepted the residual loss. With six beliefs,
AquaMend Myopic, depth-2 lookahead, and depth-3 lookahead all obtained an expected risk of $4.000$, whereas Exact Adaptive evaluated the complete chain
and reduced risk to $3.750$. Thus, the near tie between AquaMend and depth-2 lookahead on T4-Stress does not imply that finite-depth and complete adaptive
planning are generally equivalent.

The improvement from exact planning was accompanied by greater computational cost. With six beliefs, depth-2 lookahead required a median of $30.483$ ms and
$726.8$ KiB, while Exact Adaptive required $54.056$ ms and $1487.6$ KiB. Exact Adaptive is therefore used to characterize the value and computational
cost of complete planning, rather than as a replacement for the executable myopic policy.

\paragraph{Scope.}
M1-Depth is a controlled mechanism diagnostic and does not replace standard T4, T4-Stress, or the random-DAG M1 benchmark. It assumes exactly one changed belief
and independent, fully resolving probes. The results therefore establish the value of exact multi-step planning only within this verifiable subclass, and
the performance of Exact Adaptive is not attributed to AquaMend.

\section{Physics-grounded diagnostic results}
\label{app:physical_diagnostics}

AquaMend achieved attribution macro precision/recall of $0.994/0.952$ when averaged over labels with nonzero test support. Under a strict four-label macro
average that assigns zero to the unsupported \textsc{unknown} class, the values are $0.745/0.714$. The test split therefore does not establish performance on
the \textsc{unknown} category.

We separately evaluated 40 T2 pre-commit safety episodes per policy. All four policies obtained 40/40 safety successes, indicating that the shared safety
mechanism was effective but providing no comparative advantage for AquaMend.
\section{Earlier experimental evaluations}
\label{app:legacy}
\label{app:earlier_experiments}
These earlier T4, T4-Stress, MuJoCo, and M1 evaluations are retained from the supplied manuscript under their original protocols and cost definitions.
They are separate from the completed 32-scenario physical evaluation and the eight-scenario Q1 supplement.
Their sample counts, proxy losses, baseline implementations, and significance families are not pooled with those of the main evaluation.
The numerical records are retained without recomputation; independent verification requires the original code and raw records.
Oracle rows are reference policies, subject to the bound conditions in Appendix~\ref{app:prop2}.
The earlier restart-normalized saving scores below must not be read as the complete operational-cost metric defined for the current main batch.

\subsection{Standard T4 results}
\begin{table}[!htbp]
\setlength{\belowcaptionskip}{10pt}
\revisioncolor
\caption{Main comparison on T4 (reversible regime, both graph granularities).
Each method is evaluated on 4,900 episodes. Brackets report 95\% bootstrap confidence intervals for $\mathbb{E}[\mathcal{L}]$. The final column reports Holm--Bonferroni-adjusted paired Wilcoxon $p$-values against AquaMend.
ZRR is evaluated on false-alarm episodes.}
\label{tab:main}
\begin{center}
\resizebox{\linewidth}{!}{
\begin{tabular}{lccccccc}
\toprule
\textbf{Method} & $\mathrm{Succ}\uparrow$
& $\mathbb{E}[\mathcal{L}]\downarrow$
& $\mathrm{Save}_C\uparrow$
& $\#P$
& $\#R$
& $\mathrm{ZRR}\uparrow$
& $p$ (Holm) \\
\midrule
Oracle (reference)
& 1.000 & 3.31 [3.21, 3.40] & 0.768
& 0.00 & 1.22 & 1.00 & $<10^{-3}$ \\
Restart
& 0.799 & 19.52 [19.23, 19.83] & $-0.370$
& 3.00 & 3.00 & 0.00 & $<10^{-3}$ \\
Blind retry
& 0.252 & 23.70 [23.26, 24.10] & $-0.663$
& 2.00 & 0.00 & 1.00 & $<10^{-3}$ \\
VeriTrace (rollback only)
& 0.204 & 39.64 [39.05, 40.21] & $-1.782$
& 0.00 & 2.50 & 0.00 & $<10^{-3}$ \\
Troubleshooting
& 0.555 & 20.26 [19.69, 20.76] & $-0.422$
& 1.00 & 1.50 & 1.00 & $<10^{-3}$ \\
Belief-space replan
& 0.204 & 33.60 [32.99, 34.18] & $-1.358$
& 0.00 & 0.26 & 1.00 & $<10^{-3}$ \\
TMS revision
& 0.138 & 32.89 [32.29, 33.45] & $-1.308$
& 0.00 & 0.00 & 1.00 & $<10^{-3}$ \\
Decoupled/greedy/lookahead
& 0.999 & 6.09 [5.97, 6.21] & 0.572
& 2.29 & 1.22 & 1.00 & 1.000 \\
\textbf{AquaMend (ours)}
& \textbf{0.999} & \textbf{6.09 [5.97, 6.21]} & \textbf{0.572}
& \textbf{2.29} & \textbf{1.22} & \textbf{1.00} & -- \\
\bottomrule
\end{tabular}
}
\end{center}
\end{table}

On standard T4, AquaMend achieved a corrected-success rate of $0.999$ and an expected total loss of $6.09$. It substantially reduced loss relative to the traditional baselines in Table~\ref{tab:main}. However, AquaMend, decoupled recovery, greedy re-probing,
and depth-2 lookahead produced identical per-episode outcomes on this split
($p_{\mathrm{Holm}}=1.000$). Standard T4 therefore validates the executable recovery pipeline but does not distinguish the joint recovery policies. For AquaMend, the fine graph reduced $\mathbb{E}[\mathcal{L}]$ from $6.83$ to $5.36$ and the mean rollback count from $1.47$ to $0.98$, while preserving
$\mathrm{Succ}=0.999$ and $\mathrm{ZRR}=1.00$.
On the separately evaluated 1,200 irreversible T4 episodes, AquaMend, decoupled recovery, greedy re-probing, and depth-2 lookahead each obtained
$\mathrm{Succ}=0.500$ and $\mathbb{E}[\mathcal{L}]=2.73$, with an escalation rate of $0.50$. The oracle obtained the same success rate with
$\mathbb{E}[\mathcal{L}]=1.12$. When correction requires undoing an already-executed irreversible action, recovery escalates rather than assigning a finite rollback cost.

\subsection{Joint recovery under T4-Stress}

The earlier T4-Stress benchmark comprised common-cause failures, shared downstream closures, budgeted sensing, and held-out combinations of these factors. Rank ties and adversarial evidence conflicts were treated as boundary diagnostics and were not averaged into the primary claim. Each method was evaluated on the same 4,600
paired episodes across five seeds.

\begin{table}[!htbp]
\setlength{\belowcaptionskip}{10pt}
\revisioncolor
\caption{Main comparison on the T4-Stress earlier benchmark. Each method is evaluated on 4,600 frozen paired episodes. The final column reports two-sided paired Wilcoxon $p$-values against AquaMend; Holm correction is applied separately to the four designated non-oracle comparisons and the five supplemental traditional-baseline comparisons.}
\label{tab:stress}
\begin{center}
\resizebox{\linewidth}{!}{%
\begin{tabular}{lcccccc}
\toprule
\textbf{Method} & $\mathrm{Succ}\uparrow$ & $\mathbb{E}[\mathcal{L}]\downarrow$
& $\#P$ & $\#R$ & $\mathrm{OverBudget}\downarrow$ & $p$ (Holm) \\
\midrule
Oracle (reference)      & 1.000 & 7.83  & 0.00 & 2.98 & 0.000 & -- \\
\textbf{AquaMend (ours)} & \textbf{0.976} & \textbf{12.47} & 2.88 & 2.97 & 0.001 & -- \\
Lookahead                 & 0.976 & 12.47 & 2.88 & 2.97 & 0.001 & 0.317 \\
Decoupled                 & 0.575 & 47.83 & 3.61 & 1.83 & 0.403 & $<10^{-3}$ \\
Greedy re-probe           & 0.575 & 46.14 & 3.61 & 1.83 & 0.403 & $<10^{-3}$ \\
Restart                   & 0.333 & 53.16 & 5.74 & 3.48 & 0.000 & $<10^{-3}$ \\
Blind retry               & 0.000 & 45.71 & 2.00 & 0.00 & 0.000 & $<10^{-3}$ \\
VeriTrace                 & 0.000 & 63.48 & 0.00 & 2.98 & 0.000 & $<10^{-3}$ \\
Troubleshooting           & 0.172 & 32.93 & 1.00 & 3.08 & 0.000 & $<10^{-3}$ \\
Belief-space replan       & 0.000 & 55.65 & 0.00 & 0.00 & 0.000 & $<10^{-3}$ \\
TMS revision              & 0.000 & 55.65 & 0.00 & 0.00 & 0.000 & $<10^{-3}$ \\
\bottomrule
\end{tabular}%
}
\end{center}
\end{table}

T4-Stress separated the policies that tied on standard T4. AquaMend achieved
$\mathrm{Succ}=0.976$ and $\mathbb{E}[\mathcal{L}]=12.47$, compared with $0.575/47.83$ for decoupled recovery, $0.575/46.14$ for greedy re-probing, and $0.333/53.16$ for restart. The paired mean loss reductions were $35.36$, $33.67$, and $40.69$, respectively, and all remained significant after Holm correction ($p<10^{-3}$). AquaMend also reduced the budget-exceeded rate to $0.001$, compared with $0.403$ for both decoupled and greedy recovery. Its loss was also lower than each supplemental traditional baseline ($p<10^{-3}$).

Depth-2 lookahead matched AquaMend to the reported precision on T4-Stress. Their paired mean loss difference was $3.3\times10^{-5}$, with 4,599 ties among 4,600 episodes and $p_{\mathrm{Holm}}=0.317$. Thus, these experiments support an advantage over decoupled, greedy, restart, and the traditional baselines, but not over depth-2 lookahead. The reported diagnostics were consistent with the intended stress factors: under common-cause failures, AquaMend obtained an average loss of $10.25$ versus $79.95$ for decoupled recovery, while fine causal graphs reduced rollback cost from $10.48$ to $7.49$ relative to chain graphs.

The excluded diagnostics expose the remaining boundary. AquaMend's success rate fell to $0.025$ on hard rank ties, $0.425$ under hard adversarial evidence conflict, and $0.780$ on hard shared-closure episodes. These cases are reported as failure modes rather than included in the primary average.

\subsection{Physics-grounded execution validation}
\label{sec:physical_validation}

We further instantiated the multi-cup task in MuJoCo with particle-based fluid
simulation, contact-only grasping, and RGB-D target observation. We evaluated
four perturbation families (cup swapping, added water, sensor drift, and false
alarms) at T0 and T1 intervention times, using two severity variants and five
seeds. Each policy was therefore evaluated on the same 80 paired episodes.
The reported costs use the frozen normalized decision-cost model rather than
wall-clock runtime. A unit probe corresponds to 0.2 simulated seconds. The
restart cost was calibrated from the median duration of 28 complete successful
demonstrations (72.574 simulated seconds), giving
$C_{\mathrm{restart}}=72.574/0.2=362.87$.

\begin{table*}[t]
\centering
\caption{Physics-grounded multi-cup execution validation on the combined T0/T1
episodes. $\#R/C_R$ and $\#P/C_P$ denote the mean rollback and re-probe
counts/costs. Proxy loss is
$\mathcal{L}=C_P+C_R+\sum_k L_k$.
Restart is charged through the separately calibrated full-restart cost.
ZRR is evaluated on false-alarm episodes.}
\label{tab:physical_validation}
\vspace{10pt}
\resizebox{\textwidth}{!}{%
\begin{tabular}{lcccccc}
\toprule
\textbf{Method}
& $\mathrm{Succ}\uparrow$
& $\#R/C_R\downarrow$
& $\#P/C_P\downarrow$
& $\mathbb{E}[\mathcal{L}]\downarrow$
& $\mathrm{Save}_C\uparrow$
& $\mathrm{ZRR}\uparrow$ \\
\midrule
\textbf{AquaMend (ours)}
& 80/80
& 0.625 / 1.875
& 0.250 / 0.250
& 2.125
& 0.9941
& 1.000 \\
Decision-theoretic troubleshooting
& 80/80
& 0.750 / 2.250
& 0.075 / 0.075
& 2.325
& 0.9936
& 1.000 \\
Rollback-only
& 20/80
& 1.000 / 3.000
& 0 / 0
& 14.750
& 0.9594
& 0 \\
Restart
& 80/80
& 1.000 / 0
& 0 / 0
& 362.870
& 0
& 0 \\
\bottomrule
\end{tabular}%
}
\end{table*}

AquaMend and decision-theoretic troubleshooting both completed all 80 T0/T1 episodes. Under matched task success, AquaMend reduced the mean rollback count
from 0.750 to 0.625 and the normalized proxy loss from 2.325 to 2.125, corresponding to reductions of $16.7\%$ and $8.6\%$, respectively. The paired
loss comparison yielded a two-sided Wilcoxon $p$-value of $0.057$. These results validate the physical executability of AquaMend and indicate a favorable
cost-efficiency trend relative to troubleshooting.
\subsection{Exact small-graph validation on M1}

M1 was evaluated on 4,000 small random DAGs across five seeds. Non-oracle methods received only the graph, costs, and public posterior, while the hidden true-change set was used only to realize the final cost. The oracle alone received $\Phi^\star$. Exact Adaptive optimized the complete finite policy tree without access to the hidden realization. Blind retry, belief-space replan, and TMS revision were not included because M1 has no physical probe history, executable world-state transition model, or separately scored logical-revision transition.

\begin{table}[!htbp]
\setlength{\belowcaptionskip}{10pt}
\revisioncolor
\caption{Exact small-graph comparison on M1. Each method is evaluated on 4,000 graphs across five seeds. Realized cost is charged on the sampled hidden realization, whereas expected policy risk is computed from the public posterior before that realization is revealed. The final column reports the realized-cost gap to the oracle.}
\label{tab:m1}
\begin{center}
\resizebox{\linewidth}{!}{%
\begin{tabular}{lcccc}
\toprule
\textbf{Method} & $n$ & \textbf{Realized cost}$\downarrow$
& \textbf{Expected policy risk}$\downarrow$ & \textbf{Gap to oracle}$\downarrow$ \\
\midrule
Oracle (reference)       & 4,000 & 7.406  & --     & 0.000  \\
Exact Adaptive             & 4,000 & 8.560  & 9.253  & 1.154  \\
Lookahead (depth 2)        & 4,000 & 8.565  & 9.255  & 1.159  \\
\textbf{AquaMend (myopic)} & 4,000 & 8.594  & 9.290  & 1.188  \\
Troubleshooting            & 4,000 & 8.599  & 9.557  & 1.193  \\
Decoupled                  & 4,000 & 11.008 & 11.756 & 3.602  \\
Greedy re-probe            & 4,000 & 12.097 & 12.491 & 4.691  \\
Restart                    & 4,000 & 17.846 & 17.846 & 10.440 \\
VeriTrace                  & 4,000 & 18.323 & 18.226 & 10.917 \\
\bottomrule
\end{tabular}%
}
\end{center}
\end{table}

AquaMend obtained a mean realized cost of $8.594$ and an expected policy risk
of $9.290$. Exact Adaptive obtained $8.560/9.253$, while depth-2 lookahead
obtained $8.565/9.255$. The expected-risk ordering
$\textsc{Exact Adaptive}\leq\textsc{Lookahead}\leq\textsc{AquaMend}$ held on
all 4,000 graphs. The expected-risk difference between Exact Adaptive and AquaMend was $0.037$.
Exact Adaptive required complete policy-tree optimization and was therefore restricted to the small graphs on which exact enumeration was feasible. These results support using the myopic policy as a tractable approximation, but do not establish equivalence to or superiority over deeper planning.

M1 also quantified proxy error against exact $U$ on 4,000 independent and 4,000 correlated posterior states. Indep matched exact $U$ on the 243-state independent and rollback-disjoint subclass, but its normalized mean absolute error was $27.613\%$ over all independent states and $29.028\%$ over correlated states. Monte Carlo reduced these errors to $2.502\%$ and $1.557\%$, respectively. Beam-$K$ with $K=32$ matched exact $U$ because the current M1 support contained at most $2^5=32$ configurations; this verifies the implementation on these graphs but does not imply zero approximation error when the posterior support is larger.

\subsection{Ablation analysis}

\begin{table}[!htbp]
\setlength{\belowcaptionskip}{10pt}
\revisioncolor
\caption{Ablation results on standard T4. All rows except ``Chain vs.\ fine'' use the full 6,100-episode set, including the separately flagged irreversible subset; the graph-granularity row uses the 4,900 reversible episodes. Slash-separated values follow the order in the corresponding ``Ablation'' or
``Removed/varied'' entry. Lower $\mathbb{E}[\mathcal{L}]$ and higher $\mathrm{Succ}$ indicate better performance.}
\label{tab:ablation}
\begin{center}
\normalsize
\setlength{\tabcolsep}{4pt}
\begin{tabularx}{\linewidth}{
  @{}
  >{\raggedright\arraybackslash}p{0.15\linewidth}
  >{\raggedright\arraybackslash}p{0.17\linewidth}
  >{\raggedright\arraybackslash}X
  >{\centering\arraybackslash}p{0.13\linewidth}
  >{\centering\arraybackslash}p{0.13\linewidth}
  @{}
}
\toprule
\textbf{Ablation} & \textbf{Removed/}\newline\textbf{varied} & \textbf{Hypothesis} & $\mathbb{E}[\mathcal{L}]$ & $\mathrm{Succ}$ \\
\midrule
No re-probe        & rollback only            & re-probing is necessary to lower $\mathcal{L}$ & 38.157 & 0.213 \\
No attribution     & treat all as \external   & attribution avoids needless rollback           & 5.625 & 0.900 \\
No source separation     & self and external merged & cross-modal selection matters for drift         & 5.541 & 0.901 \\
Detect obj.\ $C_s/c$ vs.\ VoI & swap criterion & detection needs a falsification objective       & 5.431 / 10.335 & 0.901 / 0.708 \\
Chain vs.\ fine    & graph granularity        & chain benefit is zero rollback on false alarms  & 6.830 / 5.360 & 0.999 / 0.999 \\
$\tilde U$ proxy   & Indep / Beam / MC        & proxy accuracy vs.\ cost                        & 5.431 / 5.431 / 5.432 & 0.901 / 0.901 / 0.901 \\
Noisy probe        & increased observation noise & robustness to imperfect re-probes             & 9.830 & 0.734 \\
Calibration quality& good / bad ECE           & three-way choice sensitivity to calibration     & 5.431 / 5.431 & 0.901 / 0.901 \\
$L_i$ magnitude    & scale up or down         & when keep-and-ignore is selected                & 5.457 / 5.422 & 0.901 / 0.901 \\
\bottomrule
\end{tabularx}
\end{center}
\end{table}

Across the 6,100 standard-T4 ablation episodes, removing physical re-probing increased $\mathbb{E}[\mathcal{L}]$ from $5.431$ to $38.157$ and reduced
$\mathrm{Succ}$ from $0.901$ to $0.213$. Replacing the falsification-oriented detection objective with net value of information increased loss to $10.335$ and reduced success to $0.708$; increasing probe noise produced a similar degradation to $9.830/0.734$. By contrast, the posterior proxy, calibration, and $L_i$ variants changed little on standard T4. On the 1,200 common-cause T4-Stress episodes, Indep, Beam, and Monte Carlo proxies all obtained
$\mathrm{Succ}=1.000$ and $\mathbb{E}[\mathcal{L}]=10.247$ because the proxy change did not alter the selected actions; this verifies execution stability but does not establish a performance advantage for any proxy.

\begingroup

\section{Completed paired physical evaluation}
\label{app:newprotocol}

\paragraph{Design and independent units.}
The main execution-level comparison evaluates AquaMend, DTT, no extra re-probe, full restart, and the linear-chain control on the same 32 independent scene seeds.
Early T1 and late T1\_late each contain 16 scenes: four false alarms, four additions of water in A, four swaps of A/B, and four visual-sensor perturbations (A early, B late).
A separate no-$C_s$/netVoI-all ablation evaluates all candidate probes under AquaMend's joint posterior and one-step loss objective.
DTT achieved 25/32 complete recoveries and a mean complete loss of 72.116; its paired loss comparison is reported in Table~\ref{tab:physical160}.
Decision-time and candidate-count comparisons below concern the all-candidate ablation, not DTT.
The no-extra-re-probe control omits additional diagnosis; full restart executes physical restart operations rather than paying a constant proxy cost; the linear-chain control changes dependency granularity.
The reported cost, safety, and success rules were frozen before execution.
Evaluation truth is used for scoring, not supplied to the online policy.
The new batch is separate from the legacy benchmarks in Appendix~\ref{app:legacy}.

\paragraph{Physical ledger and residual losses.}
Sensing $C_P$, physical rollback $C_{\mathrm{rb}}$, and continuation $C_{\mathrm{cont}}$ are measured in simulation seconds from unique executed leaf receipts.
Shared operations and overlapping parent--child intervals are charged once; genuinely repeated executions incur new costs.
Common alarm processing, paid measurements, localization, rollback, and continued or restarted execution are included where actually performed.
The three ledger components partition these charges, so re-grasping or re-execution is not added again through an overlapping recovery aggregate.
Failed and partial executions remain in every total.
Cancelling an unexecuted planned action may cost zero; executed irreversible actions are not assigned finite rollback costs.

Residual losses are simulation-second equivalents for genuinely uncorrected task-relevant declarations: 20 for each binding declaration A/B/C, 10 for each quantity declaration A/B/C, 80 for target selection, and 40 for sensing evidence.
Quantity support uses a $\pm3$\,g tolerance and binding a 30\,mm tolerance, with the frozen safety guards unchanged.
These losses do not penalize additional rollback of correct actions and do not price every possible physical task failure.
For example, a physical execution can fail while leaving no penalized belief error, so complete success is reported alongside loss.
Stopping further diagnosis, committing to a supported physical action, and stopping the task are distinct decisions.
A safe stop is not successful recovery.

\begin{table}[!htbp]
\setlength{\belowcaptionskip}{10pt}
\revisioncolor
\caption{Mean complete-loss decomposition in the new main batch ($n=32$ per method).
Physical components are simulation seconds; residual and total losses are simulation-second equivalents.
Rounded entries can differ slightly from the rounded total.}
\label{tab:cost160}
\centering
\begin{tabularx}{\linewidth}{@{}Xrrrrr@{}}
\toprule
Method & Sensing & Rollback & Continue & Residual & Total \\
\midrule
AquaMend & 0.463 & 2.847 & 53.672 & 14.375 & 71.356 \\
DTT & 0.450 & 14.436 & 47.543 & 9.688 & 72.116 \\
No extra re-probe & 0.325 & 4.434 & 45.191 & 34.375 & 84.326 \\
Full restart & 0.781 & 24.495 & 51.315 & 14.375 & 90.966 \\
Linear chain & 0.444 & 11.899 & 40.923 & 23.125 & 76.391 \\
\bottomrule
\end{tabularx}
\end{table}

\paragraph{Operational savings and counts.}
For method $m$ and scene $n$, let $C_{n,m}=C_{P,n,m}+C_{\mathrm{rb},n,m}+C_{\mathrm{cont},n,m}$.
We report the ratio of summed operational costs,
\begin{equation}
\mathrm{Save}_C(m)=1-\frac{\sum_{n=1}^{32}C_{n,m}}{\sum_{n=1}^{32}C_{n,\mathrm{Restart}}}.
\end{equation}
This excludes residual penalties and decision wall time and is not the average of scene-level percentage savings.
AquaMend and restart have total operational costs 1823.386 and 2450.906 simulation seconds, giving 25.6\% savings.
Their complete-loss reduction is instead $19.610/90.966=21.6\%$.
The no-extra-re-probe and linear-chain controls spend less operationally than AquaMend but recover less often and have greater complete loss; reduced execution after failure is not a recovery benefit.
The probe count $\#P$ counts only additional diagnostic probes, so zero extra probes need not imply zero $C_P$.
Rollback count $\#R$ counts executed rollback leaf receipts, not abstract action nodes.
ZRR measures zero rollback on the eight predesignated false-alarm scenes confirmed by scoring truth.
ZRR-and-success requires both properties on that same denominator.

\begin{table}[!htbp]
\setlength{\belowcaptionskip}{10pt}
\revisioncolor
\caption{Supplementary physical metrics on the 32-scene main batch.
Mean-loss CI denotes the unadjusted 95\% scene-bootstrap interval for each method's mean complete loss.
ZRR is evaluated on eight predesignated false-alarm scenes and equals ZRR-and-success for the reported methods.}
\label{tab:counts160}
\centering
\begin{tabularx}{\linewidth}{@{}Xrrrrr@{}}
\toprule
Method & Mean-loss CI & $\mathrm{Save}_C$ & $\#P$ & $\#R$ & ZRR \\
\midrule
AquaMend & [64.833, 78.688] & 0.256 & 0.625 & 0.312 & 1.000 \\
DTT & [66.927, 77.555] & 0.185 & 0.563 & 1.469 & 1.000 \\
No extra re-probe & [78.399, 90.877] & 0.348 & 0.000 & 0.469 & 1.000 \\
Full restart & [83.443, 98.663] & 0.000 & 2.000 & 2.562 & 0.000 \\
Linear chain & [70.046, 83.426] & 0.305 & 0.688 & 1.281 & 1.000 \\
\bottomrule
\end{tabularx}
\end{table}

\paragraph{Paired inference.}
The four complete-loss comparisons against AquaMend comprise DTT, no extra re-probe, full restart, and the linear-chain control.
Two-sided paired Wilcoxon signed-rank tests give raw $p$-values of 0.465318, 0.116669, 0.0000202344, and 0.025822, respectively.
Holm correction across these four comparisons gives adjusted $p$-values of 0.465318, 0.233337, 0.0000809375, and 0.077465.
The independent unit is the scene; method runs and repeated timing calls do not increase that sample size.
Table~\ref{tab:physical160} reports mean paired loss differences and unadjusted 95\% paired-bootstrap confidence intervals.
Paired mean-difference intervals were computed using 2,000 bootstrap resamples of the 32 scenario pairs, sampled with replacement without stratification.
Both methods shared the same sampled scenario indices, and each resample contained 32 pairs.
The 95\% interval was defined by the 2.5th and 97.5th percentiles, using linear interpolation.
Bootstrap sampling used NumPy's PCG64 generator with seed 20260923.
Zero differences were retained, and the means and bootstrap intervals used unrounded losses.

For the signed-rank tests only, paired loss differences were rounded to nine decimal places before identifying zeros and ties.
Zero differences were excluded before ranking, and tied absolute differences received average ranks.
Two-sided $p$-values were computed by dynamic programming over the exact conditional sign-flip distribution, without normal approximation or continuity correction.
For DTT, 13 zero differences were excluded, leaving 19 nonzero pairs with no tied absolute differences and a signed-rank statistic of $W=76$.
These mean-difference intervals and the signed-rank tests summarize different aspects of the paired differences; the intervals are not adjusted for multiple comparisons.
Success counts are descriptive, with no confirmatory success-rate comparison.

The individual-method mean-loss intervals in Table~\ref{tab:counts160} were computed using 10,000 stratified percentile-bootstrap resamples, drawing four scenes with replacement within each of eight stage--condition strata, with seed 20260919.
Unrounded losses were used, and all failed executions were retained.
These intervals concern each method's mean complete loss and are distinct from the paired mean-loss-difference intervals in Table~\ref{tab:physical160}, which use 2,000 unstratified paired resamples with seed 20260923.

\begin{table}[!htbp]
\setlength{\belowcaptionskip}{10pt}
\revisioncolor
\caption{Complete online decision time in wall-clock seconds.
Each scene contributes its median of four calls; intervals describe paired scene-level mean differences and are not adjusted simultaneous intervals.}
\label{tab:timing160}
\centering
\begin{tabularx}{\linewidth}{@{}Xrrrr@{}}
\toprule
Stage & Scenes & AquaMend & All cand. & All cand.$-$AquaMend [95\% CI] \\
\midrule
All & 32 & 2.074 & 2.364 & 0.290 [0.212, 0.360] \\
Early & 16 & 1.703 & 2.362 & 0.659 [0.499, 0.795] \\
Late & 16 & 2.445 & 2.366 & $-$0.080 [$-$0.106, $-$0.051] \\
\bottomrule
\end{tabularx}
\end{table}

\paragraph{Decision timing and coverage.}
Each method contributes the median of four complete \texttt{next\_step} calls per scene; these medians are averaged over the 32 scenes (16 per stage).
Measurements use the same hardware, cold single-round caches, and rotating order after physical execution.
FIT-file loading and input restoration are outside the clock, as are physical execution and file I/O.
Alarm conditioning, belief refresh, $C_s$/candidate computation, and actual command construction are inside the measured call.
Both methods use shared caching, and the all-candidate ablation does not compute unused $C_s$ scores.
The overall 12.3\% reduction therefore does not imply a physical-execution speedup or uniform real-time performance.

Table~\ref{tab:screen160} counts all actual trajectory decisions, including those following different observations or actions.
It is a mechanism summary, not a matched timing sample.
The corresponding public-initial-state calls are a separate timing record.
In the early AquaMend trajectories, screening evaluates 48 of 96 enumerated candidates; in the late trajectories all 96 are evaluated because screening is uncovered.
This observed reduction does not prove that screening preserves the optimal action outside the tested states.

\begin{table}[!htbp]
\setlength{\belowcaptionskip}{10pt}
\revisioncolor
\caption{Actual-trajectory candidate counts in the new main batch.
Elig.: screening-eligible; used: screening applied; uncov.: uncovered fallback; reduced: decisions reducing candidates; enum./eval.: candidate totals; branches: predictive branches.
Restart has no such decision calls; no-extra-re-probe has 16 decisions per stage and zero candidate evaluations.}
\label{tab:screen160}
\centering
\begin{tabular}{@{}llrrrrrrrr@{}}
\toprule
Method & Stage & Calls & Elig. & Used & Uncov. & Reduced & Enum. & Eval. & Branches \\
\midrule
AquaMend & Early & 25 & 24 & 24 & 0 & 20 & 96 & 48 & 7383 \\
AquaMend & Late & 27 & 24 & 0 & 24 & 0 & 96 & 96 & 11094 \\
All cand. & Early & 25 & 0 & 0 & 0 & 0 & 96 & 96 & 11071 \\
All cand. & Late & 27 & 0 & 0 & 0 & 0 & 96 & 96 & 11094 \\
Chain & Early & 35 & 31 & 31 & 0 & 26 & 110 & 52 & 7692 \\
Chain & Late & 30 & 28 & 0 & 28 & 0 & 102 & 102 & 11220 \\
\bottomrule
\end{tabular}
\end{table}

\paragraph{Failures remain in the comparison.}
AquaMend and the all-candidate ablation fail on the same four scenes (Table~\ref{tab:fail160}).
Across all 32 scenes, the numbers of safe stops are 3, 3, 8, 3, and 7 for AquaMend, all candidates, no extra re-probe, restart, and linear chain, respectively.
The no-extra-re-probe control pours successfully in 22 scenes but completes recovery in 21; scene \texttt{s23} retains incorrect quantity and sensing declarations despite successful pouring.
All five implementations in the original report pass its physical-quality check on all scenes, which is not the complete recovery criterion.
The retained per-execution data include all unsuccessful runs, not replacements or success-only costs.

\begin{table}[!htbp]
\setlength{\belowcaptionskip}{10pt}
\revisioncolor
\caption{The four unsuccessful AquaMend scenes, also shared by the all-candidate ablation.
Residual loss is charged only to the listed uncorrected declarations.
``Not recorded'' does not infer a cause from task failure.}
\label{tab:fail160}
\centering
\begin{tabularx}{\linewidth}{@{}l>{\raggedright\arraybackslash}Xlrr@{}}
\toprule
Scene & Residual declarations & Safe stop & Residual & Complete loss \\
\midrule
s05 & Target & Yes & 80 & 94.450 \\
s13 & Target & Not recorded & 80 & 146.370 \\
s19 & Quantities A/B/C, target, sensing & Yes & 150 & 157.854 \\
s29 & Quantities A/B/C, target, sensing & Yes & 150 & 164.362 \\
\bottomrule
\end{tabularx}
\end{table}

\paragraph{Audit boundary.}
The author-supplied report contains all 160 main executions and all 40 Q1 executions.
The original report supplies complete-loss decompositions, 128 paired main-batch differences, 40 Q1 differences, success counts, operational totals, and original-family signed-rank/Holm results.
The DTT results and paired loss inference are reported separately from the archived all-candidate ablation records.
The supplied confidence intervals and timing summaries have not been independently reproduced from raw calls.
Raw physical receipts, FIT records, the frozen implementation, and the files named by the report's hashes remain necessary for an independent reproducibility check.
The author-supplied screening description is summarized in Appendix~\ref{app:screening}; independent reproduction also requires the frozen candidate-generation and observation-kernel implementation.
The earlier experimental batches require their own source verification; preregistration, calibration, and human-auditing claims cannot be inferred from the current report.
The manuscript revision does not constitute an independent simulator rerun or verification of data separation.

\section{Separate Q1 posterior-sensitivity supplement}
\label{app:q1}

\paragraph{Finite joint posterior and declaration versions.}
The FIT model has 72 joint configurations associated with 194 records in 11 groups.
Relative physical changes and persistent sensor errors are represented jointly, with predictions, $C_s$, declaration-invalidity queries, and candidate losses conditioned on the same evidence history.
The alarm is incorporated once, and old pre-probe declarations are distinguished from refreshed current declarations.
Correctly identifying an unreliable source does not entail that a usable healthy measurement exists.
Coverage of $C_s$ screening is distinct from empirical probability calibration.
The exact coverage predicate and screening rule require the frozen code specification; no score-to-Platt or conformal procedure is inferred for this implementation.

\paragraph{Five fixed posterior settings.}
Eight new scene seeds were executed under five AquaMend settings, giving 40 executions separate from the 160-run batch.
The settings, seeds, costs, budgets, controls, and success criterion were fixed before execution.
After the common alarm posterior $p(s)$, one perturbation is applied:
\begin{align}
 p_{\mathrm{original}}(s)&=p(s), &
 p_{\mathrm{concentrated}}(s)&\propto p(s)^2, &
 p_{\mathrm{dispersed}}(s)&\propto p(s)^{1/2},\notag\\
 p_{\mathrm{negative}}(s)&\propto p(s)e^{-(\ln4)B(s)}, &
 p_{\mathrm{positive}}(s)&\propto p(s)e^{+(\ln4)B(s)},
\end{align}
where $B(s)$ indicates that at least one known old declaration is invalid in model configuration $s$.
This indicator comes from the model, not scoring truth, and is distinct from rollback cost $B_i$.
Each distribution is normalized, then subsequent observations follow ordinary Bayesian updates with the original kernel; marginal probabilities, predictions, $C_s$, and losses are recomputed.
Zero-probability states remain outside support, so these transformations cannot test missing latent configurations.

\begin{table}[!htbp]
\setlength{\belowcaptionskip}{10pt}
\revisioncolor
\caption{Closed-loop Q1 results on eight new paired scenes, AquaMend only.
Every setting has complete success and pour success of 7/8.
Losses are simulation-second equivalents; differences are setting minus original, with exploratory 95\% scene-paired bootstrap intervals.}
\label{tab:q1closed}
\centering
\begin{tabularx}{\linewidth}{@{}Xrrrrr@{}}
\toprule
Setting & Sensing & Rollback & Continue & Mean loss & Difference [95\% CI] \\
\midrule
Original & 0.475 & 2.828 & 50.611 & 63.914 & 0.000 [0.000, 0.000] \\
Concentrated & 0.475 & 2.828 & 50.611 & 63.914 & 0.000 [0.000, 0.000] \\
Dispersed & 0.525 & 2.863 & 50.711 & 64.099 & 0.186 [0.023, 0.441] \\
Negative bias & 0.475 & 2.828 & 50.611 & 63.914 & 0.000 [0.000, 0.000] \\
Positive bias & 0.550 & 3.889 & 51.525 & 65.964 & 2.051 [0.023, 5.851] \\
\bottomrule
\end{tabularx}
\end{table}

\paragraph{Realized costs and concentrated sensitivity.}
All five settings fail safely on \texttt{q01}, with residual loss 80; the other seven scenes succeed.
Mean residual loss is therefore 10 for every setting in Table~\ref{tab:q1closed}.
The complete loss on \texttt{q01} is 88.126 under dispersion and 87.844 under each other setting.
Dispersion changes realized cost on \texttt{q00}, \texttt{q01}, and \texttt{q04} by 1.022, 0.282, and 0.182, respectively.
Positive bias changes \texttt{q00}, \texttt{q04}, and \texttt{q07} by 1.022, 0.182, and 15.202.
The last contributes 92.7\% of the total 16.406 increase and remains in the analysis.
Concentration and negative bias preserve each scene's realized complete loss in this batch.
Unchanged success counts therefore do not imply unchanged cost, and equal costs on eight scenes do not establish broad robustness.

Q1 intervals are exploratory ordinary paired scene-bootstrap percentile intervals with 10,000 draws and seed 20260920.
There is one scene per stage--condition cell, so within-cell variance is not meaningfully estimated by stratified resampling.
We make no confirmatory sensitivity significance claim or model selection from these intervals.
The independent unit is the scene ($n=8$), not the 40 method-setting executions.

\paragraph{Post-hoc probability evaluation.}
The probability analysis uses the 32 public alarm states from the main TEST batch, with no refitting or TEST-label model selection.
It is post hoc and does not create 32 additional physical experiments or retroactively registered endpoints.
Queries predict invalidity of a specified declaration version, not task success or support feasibility.
Tables~\ref{tab:q1old} and~\ref{tab:q1current} report Brier scores and ECE with ten fixed bins.
Each row has 32 scenes, zero missing truth, zero unknown support, and 16 states outside $C_s$ coverage.
Declarations within a scene are not independent samples; category events are overlapping unions of invalid propositions, not mutually exclusive causes.
All-negative categories can have zero Brier score without establishing detection power for absent faults.

\begin{table}[!htbp]
\setlength{\belowcaptionskip}{10pt}
\revisioncolor
\caption{Old declarations before refresh: descriptive invalidity-probability metrics on 32 main TEST scenes.}
\label{tab:q1old}
\centering
\begin{tabularx}{\linewidth}{@{}Xrrrr@{}}
\toprule
Event & Invalid rate & Mean $\hat q$ & Brier & ECE \\
\midrule
binding\_A & 0.2500 & 0.2500 & 0.0000 & 0.0000 \\
binding\_B & 0.2500 & 0.2500 & 0.0000 & 0.0000 \\
binding\_C & 0 & 0.0000 & 0.0000 & 0.0000 \\
rank\_A & 0.2500 & 0.2030 & 0.1143 & 0.1085 \\
rank\_B & 0 & 0.0622 & 0.0310 & 0.0622 \\
rank\_C & 0 & 0.0000 & 0.0000 & 0.0000 \\
min\_target & 0.1562 & 0.1532 & 0.0736 & 0.0582 \\
sensor\_valid & 0.2500 & 0.2652 & 0.1453 & 0.0756 \\
Union: binding & 0.2500 & 0.2500 & 0.0000 & 0.0000 \\
Union: quantity & 0.2500 & 0.2652 & 0.1453 & 0.0756 \\
Union: target & 0.1562 & 0.1532 & 0.0736 & 0.0582 \\
Union: sensor\_health & 0.2500 & 0.2652 & 0.1453 & 0.0756 \\
\bottomrule
\end{tabularx}
\end{table}

\begin{table}[!htbp]
\setlength{\belowcaptionskip}{10pt}
\revisioncolor
\caption{Current declarations at the first refresh: descriptive invalidity-probability metrics on 32 main TEST scenes.}
\label{tab:q1current}
\centering
\begin{tabularx}{\linewidth}{@{}Xrrrr@{}}
\toprule
Event & Invalid rate & Mean $\hat q$ & Brier & ECE \\
\midrule
binding\_A & 0 & 0.0000 & 0.0000 & 0.0000 \\
binding\_B & 0 & 0.0000 & 0.0000 & 0.0000 \\
binding\_C & 0 & 0.0000 & 0.0000 & 0.0000 \\
rank\_A & 0.1250 & 0.2205 & 0.1106 & 0.0955 \\
rank\_B & 0.1250 & 0.0788 & 0.0228 & 0.0462 \\
rank\_C & 0 & 0.0000 & 0.0000 & 0.0000 \\
min\_target & 0.1562 & 0.1583 & 0.0736 & 0.0582 \\
sensor\_valid & 0.1875 & 0.2008 & 0.1334 & 0.0738 \\
Union: binding & 0 & 0.0000 & 0.0000 & 0.0000 \\
Union: quantity & 0.2500 & 0.2993 & 0.1334 & 0.0758 \\
Union: target & 0.1562 & 0.1583 & 0.0736 & 0.0582 \\
Union: sensor\_health & 0.1875 & 0.2008 & 0.1334 & 0.0738 \\
\bottomrule
\end{tabularx}
\end{table}

Calibration of later post-action or post-pour declarations is unavailable without contemporaneous saved posteriors and truth.
Causal attribution of persistent sensor-error profiles is also unavailable because matched latent-profile truth is absent.
Missing class-conditional denominators are reported as unavailable, not zero, and success is not used to impute missing truth.

\begin{table}[!htbp]
\setlength{\belowcaptionskip}{10pt}
\revisioncolor
\caption{Offline first-command sensitivity at 32 main TEST alarm states.
Predicted loss is a model expectation, not realized physical loss.
These states and denominators differ from the eight-scene Q1 closed-loop supplement.}
\label{tab:q1offline}
\centering
\begin{tabularx}{\linewidth}{@{}Xrrr@{}}
\toprule
Setting & Changed commands & Mean posterior TV & Mean predicted loss \\
\midrule
Original & 0/32 & 0.0000 & 63.9249 \\
Concentrated & 1/32 & 0.0516 & 63.7192 \\
Dispersed & 8/32 & 0.0644 & 64.4001 \\
Negative bias & 0/32 & 0.1450 & 62.5118 \\
Positive bias & 8/32 & 0.1446 & 65.2785 \\
\bottomrule
\end{tabularx}
\end{table}

The offline command changes in Table~\ref{tab:q1offline} are not counterfactual rollouts.
After commands diverge, archived trajectories are not reused to assign realized costs to a different policy.
Only the separately executed Q1 trajectories support the realized sensitivity results in Table~\ref{tab:q1closed}.

\endgroup

\rev{\section{Screening and declaration implementation scope}
\label{app:screening}
\paragraph{Coverage and candidate retention.}
The supplied implementation description specifies screening coverage at the early T1 stage, with at least two relevant FIT history groups and prototype distance at most 6 under the code's 1\,g and 5\,mm normalization scales.
Screening is applied only when all legal candidates satisfy coverage; otherwise the legal set is retained in full.
For each declaration awaiting detection, candidates are ranked by conditional power per probe cost, and their selected candidates are combined by union.
Candidates needed after an invalidity decision, for re-estimation, or for replacing an evidence source remain available under the implementation's binding and modality conditions.
These conditions define the implemented screening domain; they do not establish posterior calibration or universally safe pruning.
The detailed normalization, tie handling, and binding/modality predicates require the frozen code specification for independent reproduction.

\paragraph{Fixed versions and support.}
For each probe evaluation, $q_b^v$ and $C_s$ refer to the same pre-probe declaration version.
A refreshed declaration receives its own value, version, and supporting evidence.
The conditional false-positive rate can be evaluated under declaration validity using the same $0.5$ rejection threshold, but no nominal significance-level guarantee is claimed.
A zero-probability conditioning event makes the corresponding conditional rate unavailable rather than zero.
Support availability and task-completion constraints are checked separately from declaration-invalidity probabilities.
}

\end{document}